\documentclass[10pt,twocolumn,letterpaper]{article}

\usepackage[pagenumbers]{cvpr} 

\usepackage{graphicx}
\usepackage{amsmath}
\usepackage{amssymb}
\usepackage{booktabs}
\usepackage{url}
\usepackage{multirow}
\usepackage{xcolor}
\definecolor{cvprblue}{rgb}{0.21,0.49,0.74}
\usepackage[pagebackref,breaklinks,colorlinks,citecolor=cvprblue]{hyperref}

\def\paperID{16710} 
\def\confName{CVPR}
\def\confYear{2024}

\title{LanGWM: Language Grounded World Model}

\author{Rudra P.K. Poudel \qquad Harit Pandya \qquad Chao Zhang\\
Cambridge Research Laboratory\\
Toshiba Europe Limited, UK\\
{\tt\small first-name.last-name@toshiba.eu}
\and
Roberto Cipolla\\
Department of Engineering\\
University of Cambridge, UK\\
{\tt\small rc10001@cam.ac.uk}
}

\begin{document}
\maketitle

\begin{abstract}
Recent advances in deep reinforcement learning have showcased its potential in tackling complex tasks. However, experiments on visual control tasks have revealed that state-of-the-art reinforcement learning models struggle with out-of-distribution generalization. Conversely, expressing higher-level concepts and global contexts is relatively easy using language. 

Building upon recent success of the large language models, our main objective is to improve the state abstraction technique in reinforcement learning by leveraging language for robust action selection. Specifically, we focus on learning language-grounded visual features to enhance the world model learning, a model-based reinforcement learning technique. 

To enforce our hypothesis explicitly, we mask out the bounding boxes of a few objects in the image observation and provide the text prompt as descriptions for these masked objects. Subsequently, we predict the masked objects along with the surrounding regions as pixel reconstruction, similar to the transformer-based masked autoencoder approach. 
 
Our proposed \textit{LanGWM: Language Grounded World Model} achieves state-of-the-art performance in out-of-distribution test at the 100K interaction steps benchmarks of iGibson point navigation tasks. Furthermore, our proposed technique of explicit language-grounded visual representation learning has the potential to improve models for human-robot interaction because our extracted visual features are language grounded.
\end{abstract}

\section{Introduction}
\label{sec:intro}

Recently, reinforcement learning (RL) models achieved state-of-the-art results on many domains. DreamerV3 \cite{dreamerv3-hafner2023} is the first algorithm that collected diamonds in the Minecraft environment without hand-designed curricula or human demonstrations. InstructGPT \cite{rlhf-instruct-gpt-ouyang2022}, a reinforcement learning from human feedback (RLHF) model significantly improved the quality of desired outputs on the foundation models. However, when state-of-the-art image-based RL control models \cite{curl-laskin20,data-aug-laskin20,dreamerv2-hafner21} are applied on out-of-distribution (OoD) generalization test in iGibson environment \cite{igibson1-shen21}, none of them performs well \cite{icwm-poudel2022}. This motivates us to rethink the RL model architecture for the image-based control task. We conjecture that all the image-based control models either need to see all the possible examples or we would need additional mechanism to deal with the OoD and sim-to-real tasks.

Conversely, language is inherently a discrete representation. This makes high level concept and global context representation easier. For example, CLIP model \cite{clip-radford2021} has shown that a chair can be detected as chair as well as furniture classes. This implies that features of semantically similar concepts reside close to each other. These are the needed ingredients for the OoD and sim-to-real generalization. Hence, in this paper we explore a representation learning architecture for RL, which improves the visual features with the help of language.

\begin{figure*}[t]
\begin{center}
\includegraphics[trim={1.0cm 4cm 4.5cm 6cm},width=1.0\linewidth]{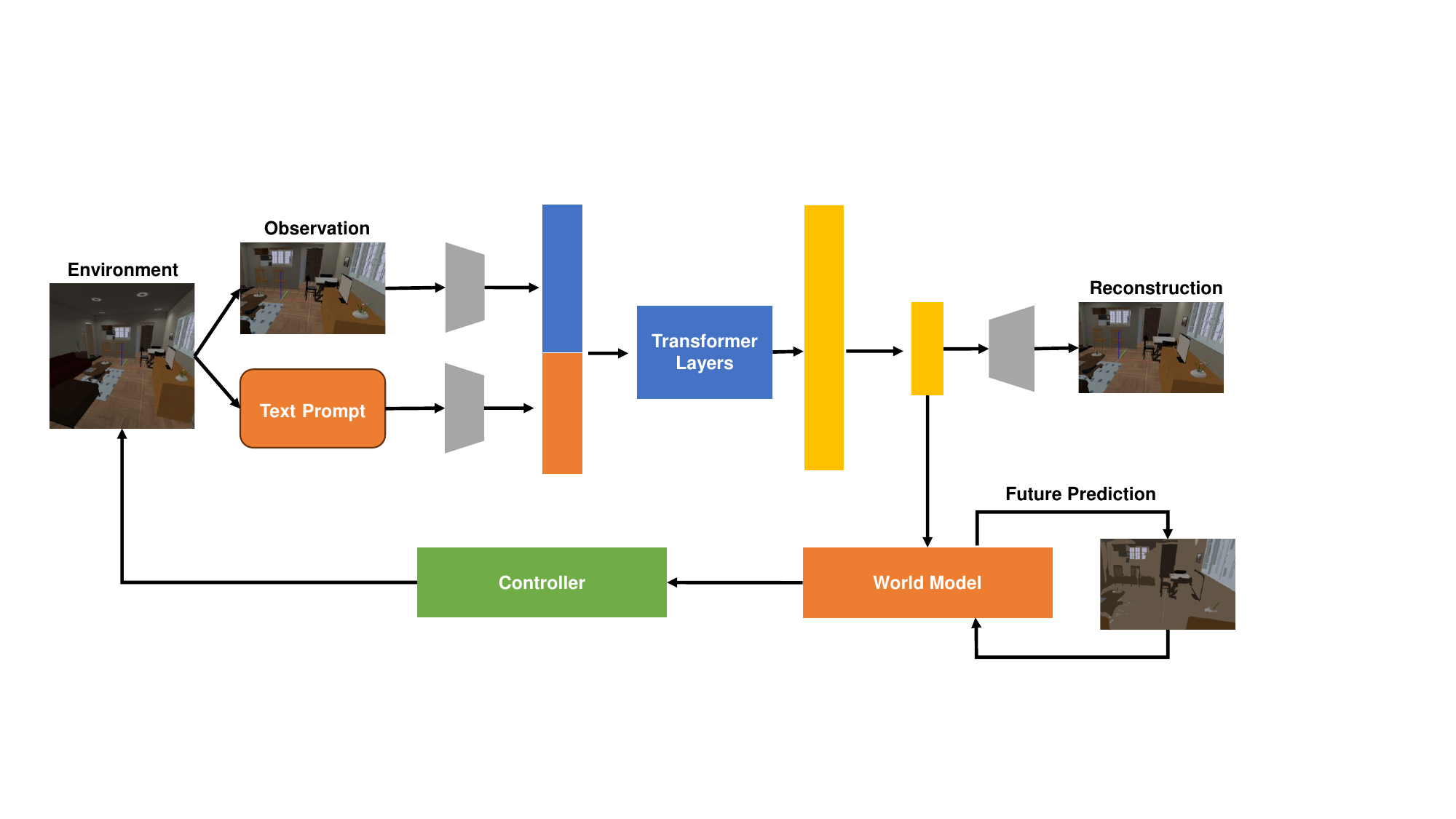}
\end{center}
\caption{Schematic diagram of proposed \textit{LanGWM: Language Grounded World Model}. LanGWM consists of three components: i) language-grounded representation learning, ii) world model, and iii) controller. Language-grounded representation learning module uses masked visual observation and complementary language prompt for task-specific features abstraction. 
World model holds the ability to predict the future states of the environment. This means the agent has the knowledge of the world its acting on.
The controller maximizes the expected rewards of the action using an actor critic approach. Since the controller is learnt separately using language grounded states of the environment, it is more robust to out-of-distribution generalization.
}
\label{fig:LanGWM}
\end{figure*}

The success of large language models (LLMs) and inherent benefit of multi-models have triggered the emergence of many vision-and-language models, like Grounding-DINO \cite{grounding-dino-liu2023}, InstructBLIP \cite{instructblip-dai2023}, SAM segment anything \cite{sam-kirillov2023} and others. However, these foundation models are resource hungry and difficult to apply on edge devices or resource limited autonomous systems. Importantly, we are motivated to know if language helps on the current state-of-the-art image-based RL control models.

RL models are generally divided into two broad categories. First, model-free RL learns the control policy end-to-end from the observations. Consequently, model-free methods \cite{data-aug-laskin20,data-aug-yarats21,curl-laskin20} often do not perform well on downstream tasks. Second, in the alternative model-based RL models also known as world model, an explicit predictive model of the world is learned, enabling the agent to plan by thinking ahead \cite{pilco-deisenroth2011,alpha-zero-silver18,wm-ha18,dreamerv2-hafner21,mwm-seo23}. The world model is learned separately from the policy, hence better suited for the generalization on the downstream tasks. In this paper, we postulate that model-based methods can be further improved using the language grounding.

To the best of our knowledge, Voltron \cite{voltron-karamcheti2023} and Dynalang \cite{dynalang-lin2023} are the closest two works related to the ours. Voltron needed pairs of images and language descriptions of their relationship. Dynalang concatenated one image frame with one word or token of the language instruction, hence their method does not scale well with the incongruent episode length and language descriptions or tokens length. Also, both methods rely on the implicit grounding of the visual features. Instead, we advocate explicit language grounding for efficient representation learning with limited data as well as limited resources setting.

In this paper, we proposed \textit{Language Grounded World Model} (LanGWM), which enforces language grounding explicitly to improve the generalization of the learned visual features. We masked the bounding boxes of the few objects in the image observation then given the language descriptions of the masked objects, we predict all the masked regions, similar to the transformer-based masked autoencoder \cite{mae-he2022}. Given these changes, we learn the navigation agent in the imagination rollouts from the learned world model with language grounded visual features, similar to the classical Dyna model \cite{dyna-sutton90} and recent Dreamer architectures \cite{dreamerv2-hafner21}, masked world model \cite{mwm-seo23} and others. Our proposed LanGWM yields state-of-the-art results on out-of-distribution test at the 100K environment and interaction steps benchmarks of the iGibson point navigation tasks. Further, our proposed explicit language grounded visual representation learning technique has potential to improve the human-robot interacts models, since visual features would be already language grounded.

In summary, unlike previous related works, our proposed LanGWM forces explicit language grounding, while also scaling well with incongurent episode length and language descriptions length. Further, our model inherits all the benefits of language and visual foundation models to the reinforcement learning settings.

In summary, our main contributions are:
\begin{enumerate}
    \item a novel explicitly language grounded visual representation learning method, which outperforms state-of-the-art models on visual control tasks.
    \item efficient alternative to visual foundation models for autonomous systems with limited resources.
\end{enumerate}

\section{Related Work}
\label{sec:related-work}

\textbf{Representation Learning in Reinforcement Learning.}
Learning reusable feature representations from large-scale unlabeled datasets has been an active research topic. In the context of computer vision, one can leverage unlabeled images and videos to learn good intermediate representations that can be used for a wide range of downstream tasks. 
Recently, VAE \cite{vae-kingma2014} has become the preferred approach for representation learning in model-based RL \cite{wm-ha18}. Since VAE does not make any additional assumptions about the downstream tasks, invariant representation learning with contrastive loss has shown promising results \cite{anand19,curl-laskin20}. Furthermore, building on the success of supervised deep learning, obtaining supervision from the data itself has been an effective approach for learning representations from the unlabelled data, known as self-supervised learning. Self-supervised learning formulates the learning task as a supervised one. In visual learning, self-supervision can be formulated using various data augmentation techniques, such as image distortion, rotation and masking \cite{ssl-dosovitskiy14,simclr-chen20,mae-he2022}. We also employ different data augmentation techniques and object masking to learn visual features, which are explained below.

\textbf{Visual-and-Language Representation Learning.}
Most previous reinforcement learning methods are focused on using raw visual input. \cite{lang-gen-temp-schwartz2020} proposed representing the state using natural language and demonstrated its effectiveness in the ViZDoom environment.
Dynalang \cite{dynalang-lin2023} unified language understanding with future prediction as a self-supervised learning objective. Once the agent learned to predict future representations in a multimodal world model, imagined model rollouts were used for action prediction. Masked Visual Pretraining (MVP) \cite{radosavovic2023real} proposed using masked auto-encoding for representation learning for per-pixel reconstruction. Reusable Representations for Robotic Manipulation (R3M) \cite{nair2022r3m} leveraged contrastive learning objectives and showed strong performance in imitation learning. However, inconsistent evaluation performance was identified following an evaluation beyond control in a study by Karamcheti \etal. Motivated by this, Voltron \cite{voltron-karamcheti2023} was proposed for language-driven representation learning. It balanced low and high-level features through language conditioning and generation. Unlike others, we explicitly force the language grounding with masked object reconstructions from the language description.

\textbf{Model-based Reinforcement Learning.}
Inspired by the fact that the human brain excels at uncovering hidden causes underlying observations, internal representations have been used to significantly influence how agents infer which actions will yield higher rewards \cite{survey-mental-simulation-hamrick2019}. Early groundwork for this concept was put forward by Sutton \cite{dyna-sutton90}, introducing the idea of using future hallucination samples rolled out from a learned world model in addition to an agent's interaction for sample-efficient learning. Moreover, 
world model planning was successfully demonstrated in the \textit{world model} by Ha \etal \cite{wm-ha18} and \textit{DreamerV2} by Hafner \etal \cite{dreamerv2-hafner21}. Recently, a popular direction involved replacing state extraction \cite{mwm-seo23} and dynamic prediction \cite{twm-robine2023} using transformer architecture, further enhancing the results \cite{mwm-seo23, twm-robine2023}. In our work, we propose learning language grounded features also utilizing the transformer architecture to improve the sample efficiency and generalization on OoD setting.

\textbf{Efficiency: Samples, Compute and Energy.}
Joint learning of auxiliary tasks with model-free RL makes them competitive with model-based RL in terms of sample efficiency. For instance, the recently proposed model-free RL method called CURL \cite{curl-laskin20} added contrastive loss as an auxiliary task and outperformed the state-of-the-art model-based RL method called Dreamer \cite{dreamer_hafner20}. Additionally, two recent works 
namely RAD \cite{data-aug-laskin20} and DrQ \cite{data-aug-yarats21}, surpassed CURL using data augmentation and no auxiliary contrastive loss. These results suggest that if an agent has access to a rich stream of data from the environments, an additional regularizer may be unnecessary. Directly optimizing the policy objective might be superior to optimizing multiple objectives. However, many complex problems lack access to rich streams of data, hence sample efficiency still matters. Finally, in many autonomous system settings, the compute and energy resources are still limited. Hence, developing overall efficient model is the focus of our work.


\section{Language Grounded World Model}
\label{sec:proposed-model}

The schematic diagram of our proposed language grounded world model is shown in Figure \ref{fig:LanGWM}. We consider the image-based visual control task as a finite-horizon, partially observable Markov decision process (POMDP). Observation space, action space and time horizon are denoted as $\mathcal{O}$, $\mathcal{A}$ and $\mathcal{T}$ respectively. The agent performs continuous actions $a_t \sim p(a_t|o_{\leq t},a_{<t})$, and receives observations and scalar rewards $o_t, r_t \sim p(o_t, r_t | o_{<t}, a_{<t})$ from the environment with unknown underline mechanisms. We use the modified transformer-based masked autoencoder \cite{mae-he2022} for the feature abstractions from observation $\Tilde{s_t} = encoder(o_t)$. The final state of the environment $s_t \sim p(s_t | s_{t-1}, a_{t-1}, \Tilde{s_t})$, predictive transition model $s_t \sim q(s_t | s_{t-1}, a_{t-1})$ and reward model $r_t \sim q(r_t | \Tilde{s_t})$ are implemented using deep neural network architectures. The goal of the DRL agent is to maximize the expected total rewards $E_p(\sum_{t=1}^{T} r_t)$. In the following sections we detail our proposed model and reasoning behind our design choice.

\subsection{LanGWM Design}
We propose a language grounded visual feature learning technique to learn the world model, which we named LanGWM: Language Grounded World Model. LanGWM consists of three main components: i) unsupervised language grounded representation learning, ii) world model, and iii) the controller and are optimized separately but sequentially. The representation learning module uses the object masking-based masked autoencoder \cite{mae-he2022}. The world model uses a recurrent neural network to output the parameters of a categorical distribution, which is used to sample the future states of the environment. The controller learns the action probability using an actor critic approach \cite{reinforce-williams92,dreamerv2-hafner21,mwm-seo23}, which maximizes the expected reward. Since the controller is learned independently of the world model, it is believed to be more suited for the downstream tasks. The three main components of the proposed LanGWM are describe below.

\subsubsection{Unsupervised language grounded representation learning}
Our main contribution lies in the language grounded feature representation learning module. We choose the masked autoencoder (MAE) \cite{mae-he2022} as our base representation learning algorithm because it is based on the scalable transformer architecture \cite{transformer-vaswani2017} and produces state-of-the-art results on unsupervised representation learning setting. Also, masking the object instances instead of random regions integrates well with the main design strategy of the masked autoencoder. Further, the world model optimizes feature learning and controller separately to learn the representative features for downstream tasks as learning from only reward predictions is expected to be poor \cite{wm-ha18}. Hence, decoder-based representation learning architectures are better suited for the world model setting.

Additionally, separating the training of the world model and controller simplifies the controller training in RL settings as most of the complexity resides in the feature extraction and the recurrent memory modules. Hence, we need a stronger feedback mechanism to learn the good features. To this end, we use monocular depth prediction; i.e. the MAE takes an RGB image as input and predicts the depth. The proposed language grounded representation learning has the following sub-modules.

\textbf{Object instance masking:} we randomly select an instance of the object then mask the smallest rectangular bounding box with the additional random margins between 0 to 10 pixels. Per RGB image observation, we select three objects at the maximum for the masking or we stop earlier if the masked region reaches 75\% of the image. In addition to the masking, we also use data augmentations to train the monocular depth prediction. We use spatial jitter, Gaussian blur, color jitter and grayscale data augmentations. Spatial jitter is implemented by first padding and then performing the random crop. Color jitter is implemented using brightness, contrast and hue augmentations in random order. All the hyperparameters of the model and data augmentations are provided in the supplementary materials.

\textbf{Language description generation:} we use language templates to generate the description of the masked object. The template is parameterised by the semantic class of the object, average horizontal position of the object center and average distance of the object regions. An example of parameterise language is: If you look \{distance\} in the \{direction\}, you will see \{object\}. All the language templates are provided in the supplementary materials.

\textbf{Object instance masked autoencoder:} the masked autoencoder (MAE) \cite{mae-he2022} is a scalable and self-supervised visual representation learning technique, which trains an autoencoder to reconstruct the randomly masked patches of the input image. MAE follows the vision transformer (ViT) \cite{vit-dosovitskiy2021} adaptation of the original transformer architecture \cite{transformer-vaswani2017} from the language domain to the vision domain. MAE converts image observation $o_t \,\epsilon \, \mathbb{R}^{H\times W \times C}$ into a sequence of $N = HW/P^2$ number of 2D patches $h_t \,\epsilon \, \mathbb{R}^{N\times(P^2C)}$, where $P$ is the patch size. Then a subset of $h_t$ patches are masked with a ratio of $m$ and the remaining patches after the masking $h_{t}^{m} \, \epsilon \, \mathbb{R}^{M\times(P^2C)}$ are then send to the ViT encoder.

\begin{gather}
\begin{aligned}
&\text{Patchify:} &&h_{t}= f^{\texttt{patch}}_{\phi}(o_{t})
\\
&\text{Masking:} &&h_{t}^{m}\sim p^{\texttt{mask}}(h_{t}^{m}\,|\,h_{t}, m)
\label{eq:mae_patchify}
\end{aligned}
\end{gather}

The ViT encoder extracts features from the remaining patches only, along with a learnable CLS token, and sends them through a series of transformer layers \cite{transformer-vaswani2017}. Then, a ViT decoder reconstructs the input by processing the tokens from the encoder and the learnable mask tokens through the similar transformer layers, followed by a single fully connected layer. The ViT decoder uses placeholder mask tokens for each masked patch.

\begin{gather}
\begin{aligned}
&\text{ViT encoder:} &&z_{t}^{m}\sim p_{\phi}(z_{t}^{m} \,|\,h_{t}^{m})
\\
&\text{ViT decoder:} &&\hat{o}_t\sim p_\phi(\hat{o}_{t} \,|\,z_{t}^{m})
\label{eq:vit}
\end{aligned}
\end{gather}

Spatial details of the scene and objects are important for the robotics domain \cite{voltron-karamcheti2023,mwm-seo23}, but small size patches are computationally expensive due to the quadratic complexity of self-attention layers in the transformer architecture. Hence, following the masked world model we have used few early convolution layer $conv$ and applied the masking in the convolutional feature maps. Since masking is applied in the features maps, we consider all the observation pixels for the reconstruction loss.

Further, we use the BERT \cite{bert-devlin2018} base uncase model with 110 million parameters to extract the feature embeds from the language descriptions $o_{t}^{l}$ of the image observation. The language description is generated from the templates. We pad or truncate the embeds to size 20 then append the one global pooled token. However, during the evaluation and test time we give constant token values of the empty description. Hence, the BERT model is only used during the model training phase and it does not affect the inference and implementation scenarios. Also, jointly learning the language embedding from the scratch would be an interesting future work.

In summary, the image  $o_{t}$ goes through a series of early convolutional layers followed by a patchify layer of patch size 1, to obtain $h^{c}_{t} \in \mathbb{R}^{N_{c} \times D}$, where $N_{c}$ is the number of convolutional features. Similarly, the language descriptions  $o_{t}^{l}$ goes through a frozen BERT model \cite{bert-devlin2018} to obtain $h^{l}_{t} \in \mathbb{R}^{N_{l} \times D}$, where $N_{l}$ is the number of language embeds. Following the MAE masking technique, $h^{c}_{t}$ is randomly masked with a ratio of $m$ to obtain $h^{c,m}_{t} \in \mathbb{R}^{M_{c} \times D}$. However, we do not mask the language embeds $h^{l}_{t}$. Then we use ViT encoder to extract the grounded visual features $z_{t}^{c,m,l}$ and decoder process $\tilde{s}_{t} \sim p^\texttt{slice-l}(z_{t}^{c,m,l})$ by slicing out the language embeds to reconstruct raw pixel values only but not the language descriptions. Our design choice is motivated by the fact that the ViT decoder can excel in decoding the masked objects, if it understands the language. Hence, the object instance masking technique is designed to explicitly force the language grounding unlike other related work \cite{voltron-karamcheti2023,dynalang-lin2023}.

Finally, language grounded feature learning has the following component,

\begin{gather}
\begin{aligned}
&\text{Obj. mask \& Lang. desc.:} &&o_{t}^{m}, o_{t}^{l}= f^{\texttt{object-mask}}(o_{t})
\\
&\text{Early convolution:} &&h^{c}_{t}= f^{\texttt{conv}}_{\phi}(o_{t}^{m})
\\
&\text{Conv tokens masking:} &&h_{t}^{c,m}\sim p^{\texttt{mask}}(h_{t}^{c,m}\,|\,h^{c}_{t}, m)
\\
&\text{Language embeds:} &&h^{l}_{t}= f^{\texttt{bert}}(o^{l}_{t})
\\
&\text{Tokens concat:} &&h_{t}^{c,m,l} \sim \texttt{concat}(h_{t}^{c,m},\,h^{l}_{t})
\\
&\text{MAE encoder:} &&z_{t}^{c,m,l}\sim p_{\phi}(z_{t}^{c,m,l} \,|\,h_{t}^{c,m,l})
\\
&\text{Env. states:} &&\tilde{s}_{t} \sim p^\texttt{slice-l}(z_{t}^{c,m,l})
\\
&\text{MAE decoder, reward:} &&\hat{o}_t,\,\hat{r}_t\sim p_\phi(\hat{o}_t,\,\hat{r}_t \,|\,\tilde{s}_{t})
\label{eq:ours_mae}
\end{aligned}
\end{gather}

We discard the concatenated language tokens after the MAE encoder and feed the visual positional tokens only $\tilde{s}_{t}$ to prevent the decoder cheating and to enforce the visual features to be grounded. Further, in order to encode the task relevant information, we also predict the reward $\hat{r}_t$ along with the observation. All autoencoder modules are optimized using the mean square loss of the depth reconstruction and reward prediction.

Further, we use depth for the reconstruction loss even though representation learning module only uses RGB observation as an input. Since depth observation is only needed during the training, our design choice does not hinder the wider application of our proposed model in cases where only an RGB sensor is available.

\subsubsection{Future predictive world model}
The representation learning model extracts what the agent sees at each time frame, but we also want our agent to remember the important events from the past. This is achieved with the memory module and implemented with a recurrent neural network. Further, the transition model learns to predict the future state using the current state and action in the latent space only, which enables future imagination without knowing the future observation since we can obtain the future action from the policy if we know the future state. Hence, this module is called a future predictive model and enables efficient latent imagination for planning \cite{wm-ha18,dreamer_hafner20}. In summary, dynamic memory and representation learning modules are tightly integrated and have the following components,

\begin{gather}
\begin{aligned}
 & \text{Representation model:} && p_\theta(s_t|s_{t-1},a_{t-1},\Tilde{s_t}) \\
 & \text{Predictive memory model:} && q_\theta(s_t|s_{t-1},a_{t-1}).
\end{aligned}
\end{gather}

Representation and future predictive KL regularizer losses were optimized jointly,

\begin{equation}
    \mathcal{L}_{WM} = E_p \left ( \sum_{t} \left ( \ell^q_t - \beta \ell^{KL}_t \right ) \right )
\end{equation}

where, $\ell^{KL}_t = KL(p(s_t|s_{t-1},a_{t-1},\Tilde{s_t}) || q(s_t|s_{t-1},a_{t-1}))$

\subsubsection{Policy Controller}
The objective of the controller is to optimize the expected rewards of the action, which is optimized using an actor critic approach. The actor critic approach considers the rewards beyond the horizon. Inspired by world model \cite{wm-ha18} and Dreamer \cite{dreamer_hafner20}, we learn an action model and a value model in the imagined latent space of the world model. The action model implements a policy that aims to predict future actions that maximizes the total expected rewards in the imagined environment. Given $H$ as the imagination horizon length, $\gamma$ the discount factor for the future rewards, action and policy model are defined as follows:
\begin{gather}
\begin{aligned}
 & \text{Action model:} && q_\phi(a_t|s_{t}) \\
 & \text{Value model:} && E_{q(\cdot|s_\tau)}{\textstyle \sum_{\tau=t}^{t+H}\gamma^{\tau-t}r_\tau}.
\end{aligned}
\end{gather}


\subsection{Implementation Details}
The proposed LanGWM expands on the publicly available code base of DreamerV2 \cite{dreamerv2-hafner21} and masked world model (MWM) \cite{mwm-seo23}. The MAE, world model and policy controller losses were optimized independently using Adam \cite{adam-kingma2014}. We have used four layers early convolution encoder with starting number of feature maps equal to 32, then doubled in every consecutive layer. The MAE parameters are provided in the supplementary materials.

To encode the task observations, we used four dense layers of size 256 with ELU activations \cite{elu-clevert2015}. The features from MAE encoder and task observation are first concatenated then average poolled using a autoencoder transformer before sending to the controller model. The detail of the average pooling module is also provided in the supplementary materials. The replay buffer capacity is set to $3e5$. We update the model parameters on every fifth interactive steps. Further, all architectural details and hyperparameters are provided in the supplementary materials. The training time of LanGWM is around 2 days on a workstation with two Nvidia GeForce RTX 3090 for 100k steps.

\section{Experiments}
\label{sec:exps}

\begin{table*}[t]
\begin{center}
\begin{tabular}{lcrrrrrr|rr}
\hline
\noalign{\smallskip}
\multicolumn{2}{c}{} & \multicolumn{2}{c}{Ihlen\_0\_int} & \multicolumn{2}{c}{Ihlen\_1\_int} & \multicolumn{2}{c}{Rs\_int} & \multicolumn{2}{c}{\textbf{Env Avg}}\\
\noalign{\smallskip}
Models & Steps & \multicolumn{1}{c}{SR} &  \multicolumn{1}{c}{SPL} & \multicolumn{1}{c}{SR} &  \multicolumn{1}{c}{SPL} & \multicolumn{1}{c}{SR} &  \multicolumn{1}{c}{SPL} & \multicolumn{1}{c}{SR} &  \multicolumn{1}{c}{SPL}\\
\noalign{\smallskip}
\hline
\noalign{\smallskip}
RAD & 100k & 0.6 & 0.01 & 0.1 & 0.00 & 0.8 & 0.01 & 0.5 & 0.01\\
CURL & 100k & 8.0 & \textbf{0.07} & 0.6 & \textbf{0.01} & 5.4 & 0.05 & 4.7 & \textbf{0.04}\\
DreamerV2 & 100k & 1.8 & 0.01 & 0.6 & 0.00 & 1.7 & 0.01 & 1.3 & 0.01\\
DreamerV2 + DA & 100k & 7.3 & 0.05 & 1.6 & \textbf{0.01} & 7.7 & 0.05 & 5.5 & \textbf{0.04}\\
MWM & 100k & 1.6 & 0.01 & 0.5 & 0.00 & 2.9 & 0.02 & 1.7 & 0.01\\
LanGWM & 100k & \textbf{8.3} & 0.05 & \textbf{2.1} & \textbf{0.01} & \textbf{9.9} & \textbf{0.06} & \textbf{6.8} & \textbf{0.04}\\
\hline
LanGWM + Obj Mask + Empty Lang & 100k & 1.6 & 0.01 & 0.5 & 0.0 & 2.3 & 0.01 & 1.5 & 0.01\\
LanGWM - Obj Mask + Empty Lang & 100k & 1.6 & 0.01 & 0.5 & 0.0 & 3.0 & 0.02 & 1.7 & 0.01\\
\color{gray} DreamerV2 + Grounding DINO & \color{gray} 100k & \color{gray} 48.9 & \color{gray} 0.45 & \color{gray} 17.0 & \color{gray} 0.14 & \color{gray} 45.4 & \color{gray} 0.38 & \color{gray} 37.1 & \color{gray} 0.33\\
\hline
\end{tabular}
\end{center}
\caption{This table shows the out-of-distribution generalization performances on iGibson 1.0 dataset for PointGoal navigation task. \textit{Success rate} (SR) and \textit{Success weighted by (normalized inverse) Path Length} (SPL) are reported. We have trained on five scenes, and tested on held-out three scenes as well as visual textures. Our proposed LanGWM outperforms state-of-the-art RL models CURL, RAD and DreamerV2 on 100k interactive steps. Even though \textit{data augmentation} (DA) improves DreamerV2, the proposed language grounded features learning technique yield even better results. LanGWM performance over the Transformer-based MWM \cite{mwm-seo23} shows the significance of the proposed language grounding technique.}
\label{tbl:igibson1-results}
\end{table*}

We evaluate the proposed LanGWM on iGibson 1.0 environment \cite{igibson1-shen21} and compare to the state of the art in model-based and model-free RL. Specifically, we use the \textit{PointGoal} navigation task from the iGibson 1.0 environment \cite{igibson1-shen21} to evaluate out-of-distribution (OoD) generalization. Following the common practice of the literature \cite{curl-laskin20,data-aug-laskin20,data-aug-yarats21,dreamerv2-hafner21}, we compare the results based on \textit{Success Rate} (SR) and \textit{Success weighted by (normalized inverse) Path Length} (SPL) at only 100k environment steps in order to test sample efficiency. In addition, we also report the results of an ablation study. State-of-the-art model-free RL methods CURL \cite{curl-laskin20} and RAD \cite{data-aug-laskin20}, and model-based RL method DreamerV2 \cite{dreamerv2-hafner21} are used for comparison. The source code used was provided by the authors. Please note that the improvement proposed by DreamerV3 \cite{dreamerv3-hafner2023} over the DreamerV2 \cite{dreamerv2-hafner21} are also complementary to our proposed LanGWM.

\subsection{Out-of-Distribution Generalization}
We have tested our proposed LanGWM on random \textit{PointGoal} navigation tasks of the iGibson 1.0 environment \cite{igibson1-shen21} for OoD generalization. The iGibson dataset contains 15 floor scenes with 108 rooms. The scenes are replicas of real-world homes with artist designed textures and materials. RGB, depth and task related observation are used for the experiments. The task related observation includes goal location, current location, and linear and angular velocities of the robot. We emphasize that image depth is only used during the training phase. The TurtleBot possible actions include rotation in radians and forward distance in meters. Following \cite{icwm-poudel2022}, we split iGibson to perform OoD generalization test; we choose five scenes for training and tested on the held-out three scenes. Also, held-out visual textures and materials for all object classes. More details are provided in the supplementary material.

In Table \ref{tbl:igibson1-results}, we report the average SR and SPL on the held-out data of three experiments with different random seeds. Our proposed LanGWM outperforms the state-of-the-art model-based RL method DreamerV2 and model-free method CURL and RAD on 100k interactive steps. Even though the original DreamerV2 implementation is not trained with \textit{data augmentation} (DA), we also included the case of DreamerV2 with DA in our evaluation to show a fair comparison. Nevertheless, the proposed LanGWM still outperforms the competitors using the language grounded visual features learning techniques. The superior performance of the LanGWM over the Transformer-based masked world model \cite{mwm-seo23} highlights the importance of language grounding for generalized visual feature learning. The MAE reconstruction loss forces to keep the spatial details, while the language description helps to learn the global context in the scene, example far vs close or a larger pixel-region as a table.

Finally, we note that even though LanGWM performance is significantly below the state-of-the-art foundation model Grounding-DINO \cite{grounding-dino-liu2023}, we still believe that our architecture is scaleable and would trigger the interest of the community on improving the visual feature with complementary properties of the language in the RL settings to reduce the simulation to real gap.

\subsection{Ablation Study}
 
The results ablating the contribution of object instance masking and language descriptions to enforce the language grounding are also shown in Table \ref{tbl:igibson1-results}.

We can see that in a reasonably complex, pixel-based control task, LanGWM is not able to learn the meaningful control without language descriptions as well as without the object instance masking. Hence, we conclude that to enforce the language grounding proposed object masking and language descriptions are both needed and complementary.

\section{Conclusion}

In this work we proposed \textit{LanGWM} the Language Grounded World Model in the reinforcement learning framework. Explicitly forcing the language grounding improves the generalization of the learned visual features and allows us to achieve state-of-the-art performance in out-of-distribution test at the 100K interaction steps benchmarks of iGibson point navigation tasks. Moreover, our method is sample efficient and has the potential to improve models for human-robot interaction as extracted visual features are language grounded.

{
    \small
    \bibliographystyle{ieeenat_fullname}
    \bibliography{main}
}

\newpage
\appendix
\onecolumn

\begin{center}{\bf {\LARGE Appendix}}
\end{center}

\section{LanGWM: Hyperparameters}
\begin{table}[htb]
\begin{minipage}{\textwidth}
\begin{center}
\begin{tabular}{lcc}
\hline
\noalign{\smallskip}
\textbf{Name} & \textbf{Symbol} & \textbf{Value} \\
\noalign{\smallskip}
\hline
\noalign{\smallskip}
Data Augmentation \\
\noalign{\smallskip}
\hline
\noalign{\smallskip}
Padding range & --- & 10\\
Hue delta & --- & 0.1\\
Brightness delta & --- & 0.4\\
Contrast delta & --- & 0.4\\
Saturation delta & --- & 0.2\\
Gaussina blur sigma min, max & --- & 0.1, 2.0\\
\noalign{\smallskip}
\hline
\noalign{\smallskip}
(Object) Masked Autoencoder \\
\noalign{\smallskip}
\hline
\noalign{\smallskip}
Dataset size (FIFO) & --- & $3\cdot10^5$ \\
iGibson visual observation size & $o$ & 128$\times$160 \\
Batch episode size & --- & 16 \\
Episode sequence length & $L$ & 50 \\
Batch size & $B$ & 16$\times$50 \\
Early conv kernels sizes & --- & 4, 4, 4, 4\\
Early conv feature maps & --- & 32, 64, 128, 256\\
Early conv strides & --- & 2, 2, 2, 2\\
Early conv activation & --- & ELU\\
Visual tokenization & --- & ($8\times$10)$\times$256\\
Language tokenization (BERT): 20 + global pool & --- & 21\\
CLS token & --- & 1\\
Encoder/decoder embed/token size/dim & --- & 256\\
Total num tokens & --- & 102\\
mask $p$ & --- & 0.75\\
Encoder/decoder heads & --- & 4\\
Encoder/decoder MLP ratio & --- & 4\\
Encoder layers & --- & 4\\
Decoder layers & --- & 3\\
Encoder/decoder layer norm epsilon & --- & $1^{-6}$\\
Encoder/decoder dropout & --- & 0.1\\
Transformer MLP activation & --- & GELU\\
Decode reward & --- & True\\
Loss & --- & MSE\\
Learning rate & --- & $1\cdot10^{-4}$ \\
Adam epsilon & $\epsilon$ & $10^{-5}$ \\
\noalign{\smallskip}
\hline
\noalign{\smallskip}
Task Observation Autoencoder \\
\noalign{\smallskip}
\hline
\noalign{\smallskip}
MLP encoder sizes & --- & 256, 256, 256, 256\\
Activation & --- & ELU\\
\noalign{\smallskip}
\hline
\noalign{\smallskip}
\end{tabular}
\end{center}
\caption{Hyperparameters of data augmentation, (object) masked autoencoder and task observation autoencoder.}
\label{tbl:recore-hparams}
\end{minipage}
\end{table}
\vfill
\clearpage

\begin{table}[htb]
\begin{minipage}{\textwidth}
\begin{center}
\begin{tabular}{lcc}
\hline
\noalign{\smallskip}
\textbf{Name} & \textbf{Symbol} & \textbf{Value} \\
\noalign{\smallskip}
\hline
\noalign{\smallskip}
Multimodel Average Feature Pooling Transformer Autoencoder \\
\noalign{\smallskip}
\hline
\noalign{\smallskip}
Encoder/decoder embed/token size/dim & --- & 256\\
Visual num tokens & --- & 81\\
Task obs num tokens & --- & 1\\
Total num tokens & --- & 82\\
Encoder/decoder heads & --- & 4\\
Encoder/decoder MLP ratio & --- & 4\\
Encoder/decoder layers & --- & 2\\
Decoder input & --- & avg(tokens)\\
Encoder/decoder layer norm epsilon & --- & $1^{-6}$\\
Encoder/decoder dropout & --- & 0.1\\
Loss & --- & MSE\\
\noalign{\smallskip}
\hline
\noalign{\smallskip}
World Model \\
\noalign{\smallskip}
\hline
\noalign{\smallskip}
Discrete latent dimensions & --- & 32 \\
Discrete latent classes & --- & 32 \\
RSSM number of units & --- & 1024 \\
KL loss scale & $\beta$ & 1.0 \\
World model learning rate & --- & $3\cdot10^{-4}$ \\
Key encoder exponential moving average & --- & 0.999 \\
\noalign{\smallskip}
\hline
\noalign{\smallskip}
Behavior: Actor-Critics \\
\noalign{\smallskip}
\hline
\noalign{\smallskip}
Imagination horizon & $H$ & 15 \\
Actor learning rate & --- & $1\cdot10^{-4}$ \\
Critic learning rate & --- & $1\cdot10^{-4}$ \\
Slow critic update interval & --- & $100$ \\
\noalign{\smallskip}
\hline
\noalign{\smallskip}
Common \\
\noalign{\smallskip}
\hline
\noalign{\smallskip}
Policy steps per gradient step & --- & 4 \\
Policy and reward MPL number of layers & --- & 4 \\
Policy and reward MPL number of units & --- & 400 \\
Gradient clipping & --- & 100 \\
Adam epsilon & $\epsilon$ & $10^{-5}$ \\
\noalign{\smallskip}
\hline
\noalign{\smallskip}
\end{tabular}
\end{center}
\caption{Hyper parameters of world model and actor-critics related modules of the proposed LanGWM.}
\label{tbl:recore-hparams}
\end{minipage}
\end{table}
\vfill
\clearpage

\section{Language Templates}
Given the horizontal-axis value of the object center (c), we randomly select one of the following template for the direction,
\begin{enumerate}
    \item Rule 1: block\_size $=$ image\_width / 4. If c $<=$ block\_size, return left; else-if c $<=$ 3$\times$block\_size, return front; else return right. 
    \item Rule 2: block\_size $=$ image\_width / 5. If c $<=$ block\_size, return  outer left; else-if c $<=$ 2$\times$block\_size, return left; else-if c $<=$ 3$\times$block\_size, return front; else-if c $<=$ 4$\times$block\_size, return right; else return outer right. 
\end{enumerate}

Language generation templates for the given object, direction and distance,

\begin{enumerate}
    \item There is \{object\} in the \{direction\} \{distance\}.
    \item A \{object\} is situated in the \{distance\} of the \{direction\}.
    \item The \{object\} is approximately \{distance\}, \{direction\} from here.
    \item The \{object\} is approximately \{distance\}, to the \{direction\}.
    \item In the \{direction\}, \{object\} is \{distance\} to me.
    \item In the image, I can see \{object\}, and it is \{distance\} \{direction\}.
    \item If you look \{distance\} in the \{direction\}, you will see \{object\}.
\end{enumerate}
\vfill
\clearpage
\section{iGibson 1.0: Train and Test Splits}
\begin{table}[htb]
\begin{minipage}{\textwidth}
\begin{center}
\begin{tabular}{lll}
\hline
\textbf{Phase} & & \textbf{Scene names} \\
\hline
Train & \hspace{1cm} & Beechwood\_0\_int, Beechwood\_1\_int, \\ 
         & \hspace{1cm} & Benevolence\_0\_int, Benevolence\_1\_int, Benevolence\_2\_int\\
         & \hspace{1cm} & Merom\_0\_int, Merom\_1\_int, \\
         & \hspace{1cm} & Pomaria\_0\_int, Pomaria\_1\_int, Pomaria\_2\_int,\\
         & \hspace{1cm} & Wainscott\_0\_int, Wainscott\_1\_int \\
         \hline
Test & \hspace{1cm} & Ihlen\_0\_int, Ihlen\_1\_int, Rs\_int\\
\hline
\end{tabular}
\end{center}
\caption{Train and test scenes split for iGibsion 1.0 environment.}
\label{tbl:igibson1-0-splits}
\end{minipage}
\end{table}

\begin{table}[htb]
\begin{minipage}{\textwidth}
\begin{center}
\begin{tabular}{lll}
\hline
\textbf{Material category} & \hspace{1.5cm} & \textbf{Held-out texture ids for test}\\
\hline
asphalt               && 06, 15             \\
bricks                && 08, 19             \\
concrete              && 06, 15, 17         \\
fabric                && 01, 02, 28         \\
fabric\_carpet        && 02, 05, 13         \\
ground                && 13, 19             \\
leather               && 03, 12             \\
marble                && 02, 03             \\
metal                 && 10, 19             \\
metal\_diamond\_plate && 04                 \\
moss                  && 01, 03             \\
paint                 && 05                 \\
paving\_stones        && 24, 38             \\
planks                && 07, 09, 16         \\
plaster               && 03                 \\
plastic               && 04, 05             \\
porcelain             && 02, 04             \\
rocks                 && 04                 \\
terrazzo              && 06, 08             \\
tiles                 && 43, 49             \\
wood                  && 02, 05, 16, 22, 32 \\
wood\_floor           && 06, 10, 17, 28     \\ 
\hline
\end{tabular}
\end{center}
\caption{iGibson 1.0 held-out texture ids for testing out-of-distribution generalization. All remaining texture ids are used during the training phase.}
\label{tbl:held-out-textures}
\end{minipage}

\end{table}
\clearpage


\begin{table}[htb]
\begin{minipage}{\textwidth}
\begin{center}
\begin{tabular}{ccccc}
\hline
\noalign{\smallskip}
\multirow{2}{*}{concrete} & \multirow{1}{*}{Train} & \includegraphics[width=0.16\textwidth,height=0.12\textwidth]{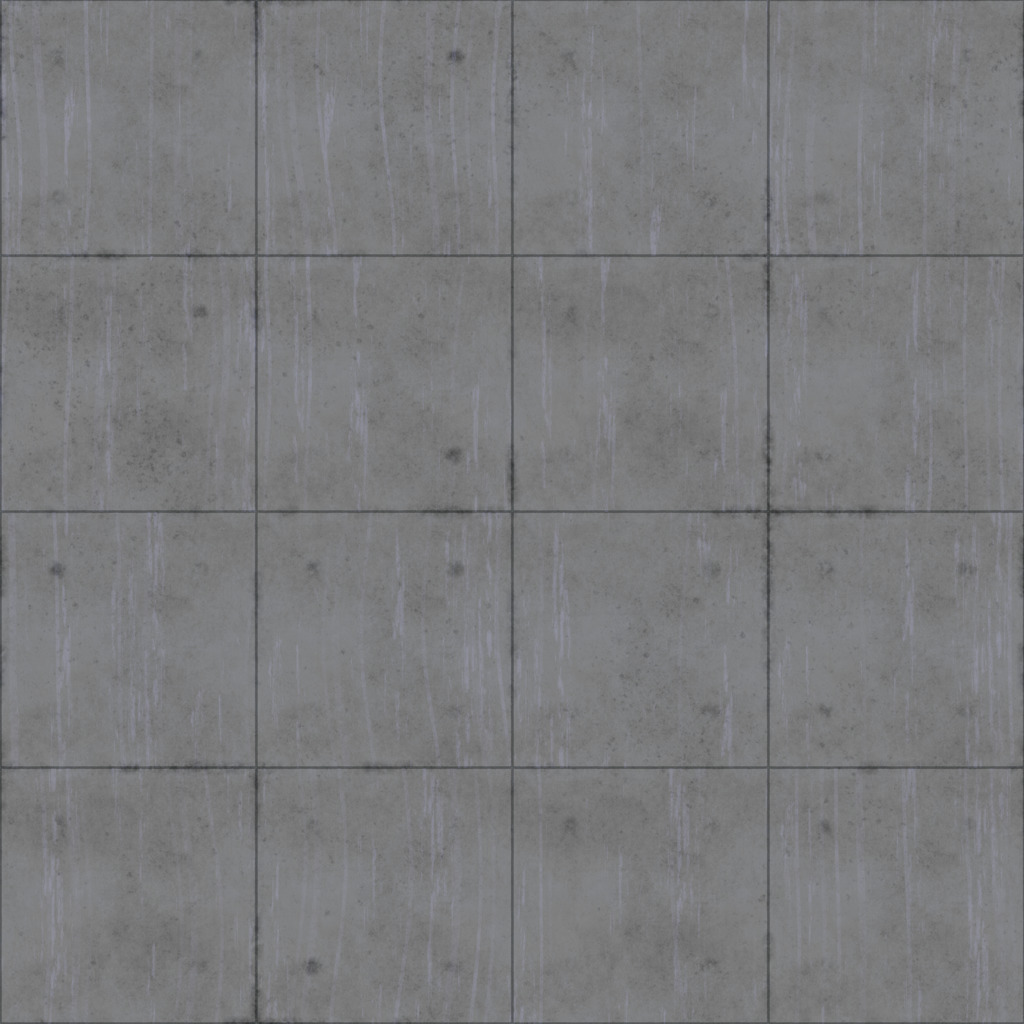} & 
                                    \includegraphics[width=0.16\textwidth,height=0.12\textwidth]{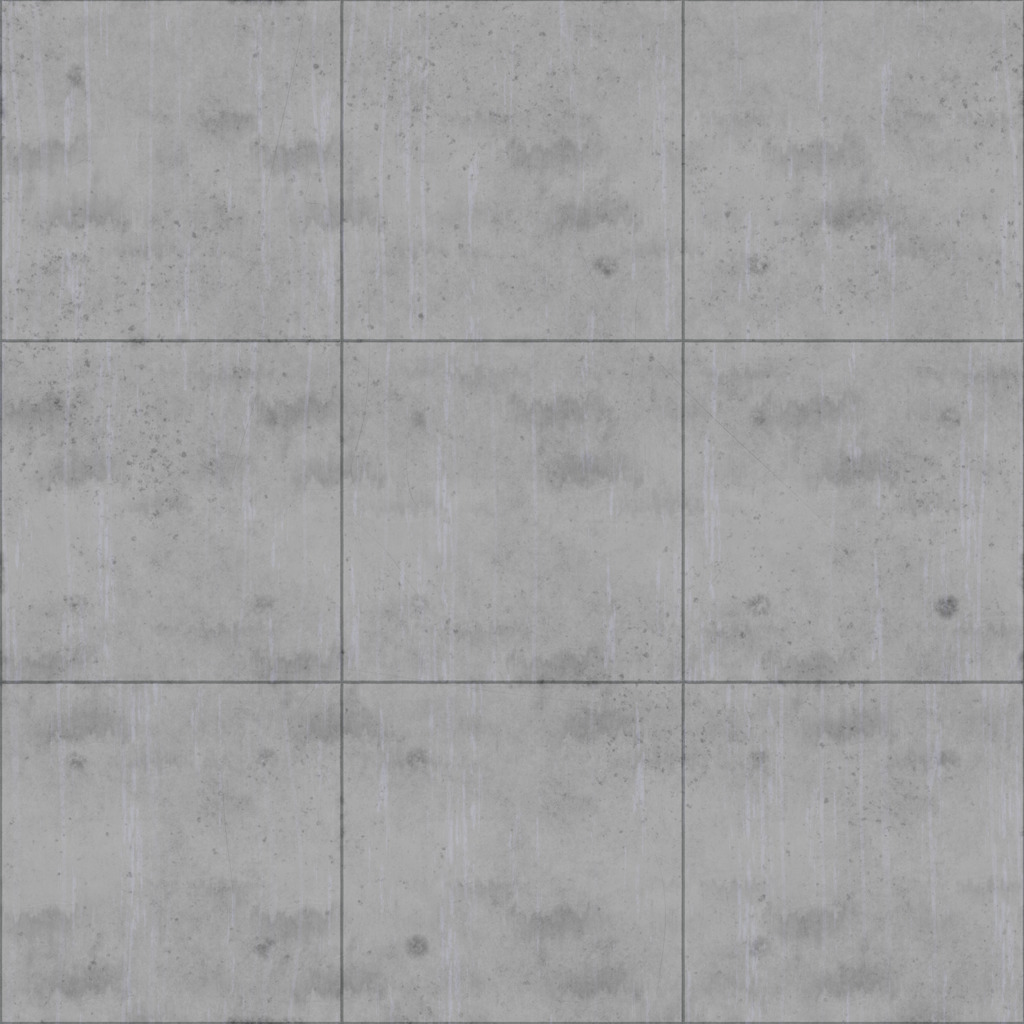} & 
                                    \includegraphics[width=0.16\textwidth,height=0.12\textwidth]{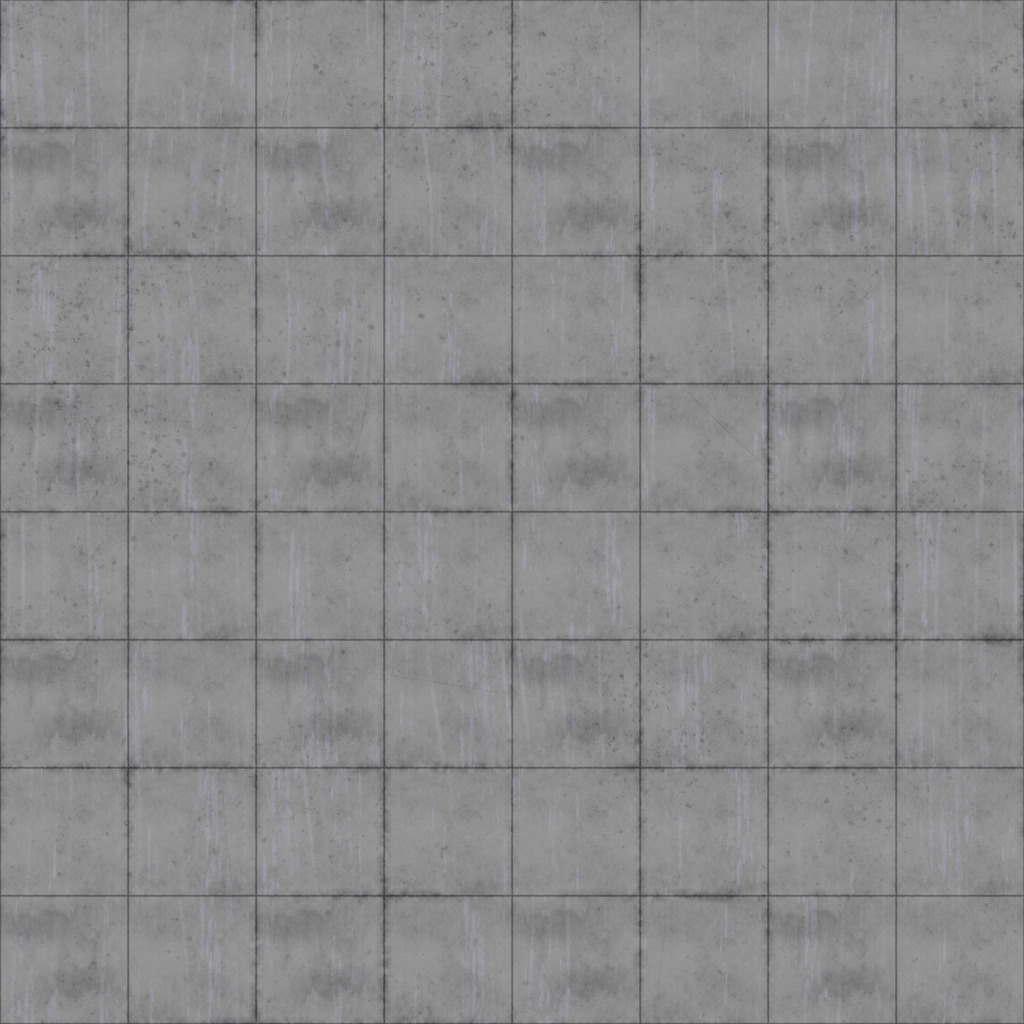} \\
                          
                          & \multirow{1}{*}{Test}  & \includegraphics[width=0.16\textwidth,height=0.12\textwidth]{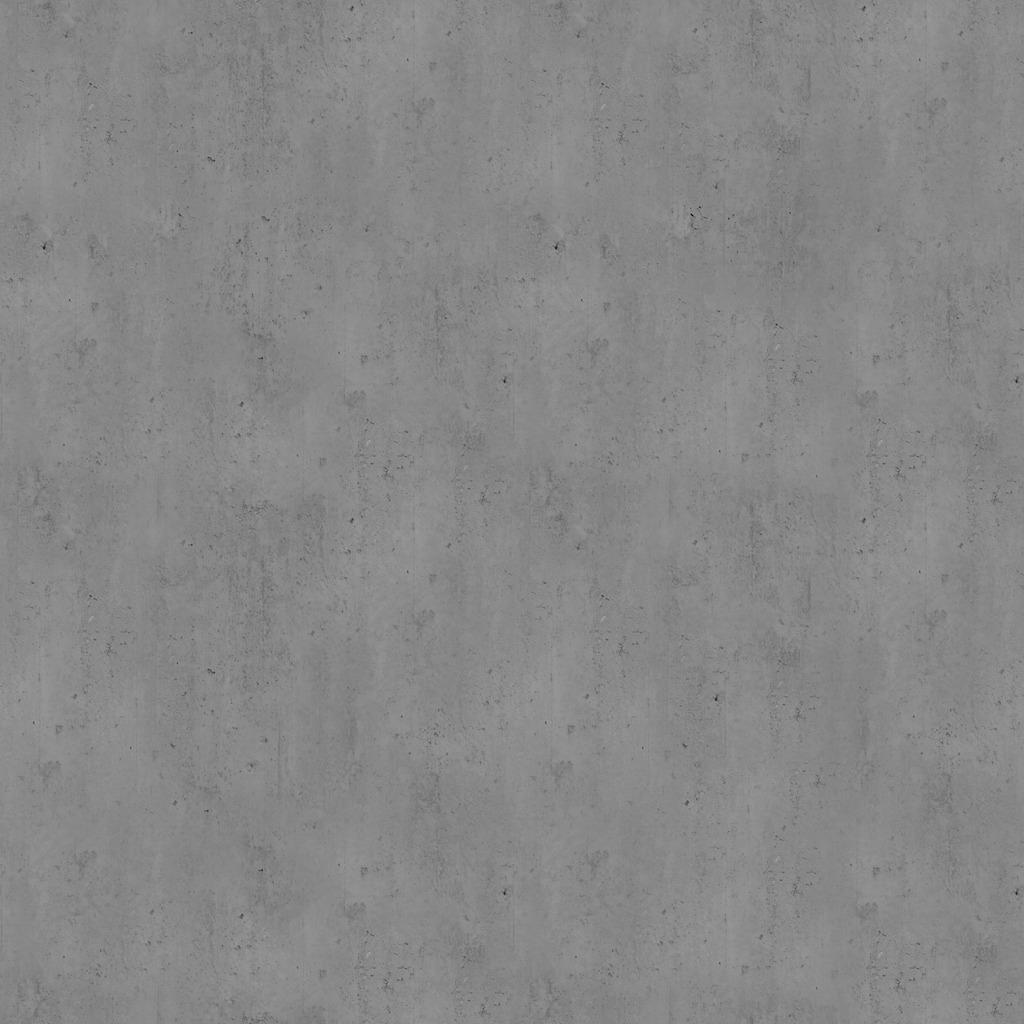} & 
                                    \includegraphics[width=0.16\textwidth,height=0.12\textwidth]{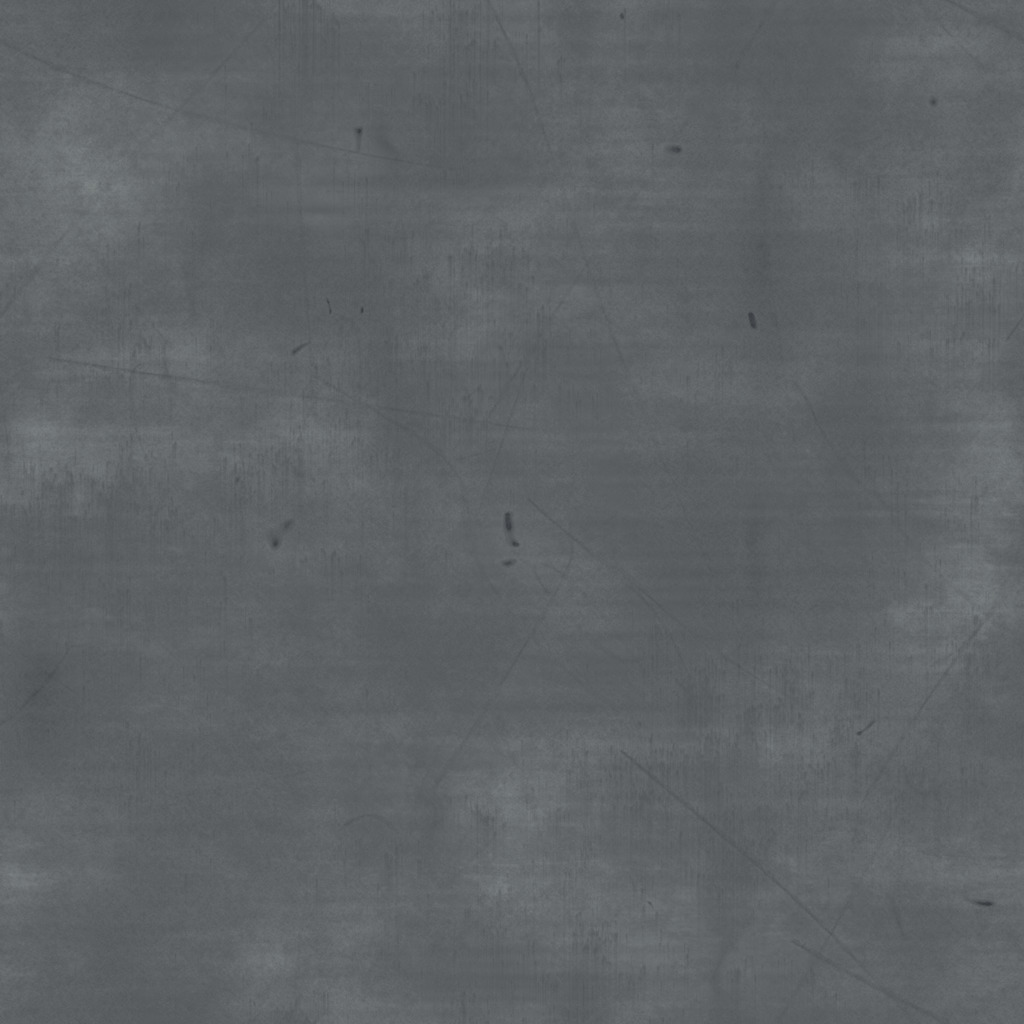} & 
                                    \includegraphics[width=0.16\textwidth,height=0.12\textwidth]{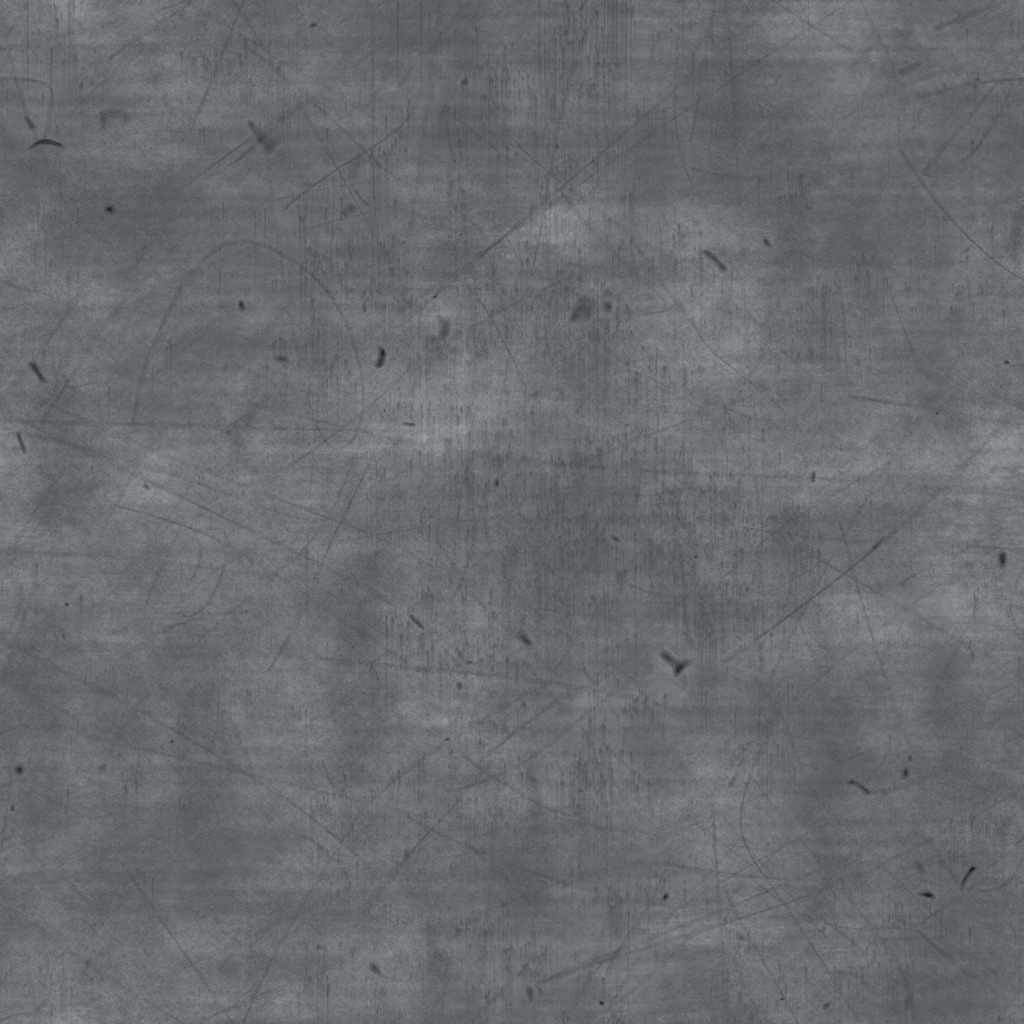} \\
\noalign{\smallskip}
\hline
\noalign{\smallskip}
\multirow{2}{*}{fabric}   & Train & \includegraphics[width=0.16\textwidth,height=0.12\textwidth]{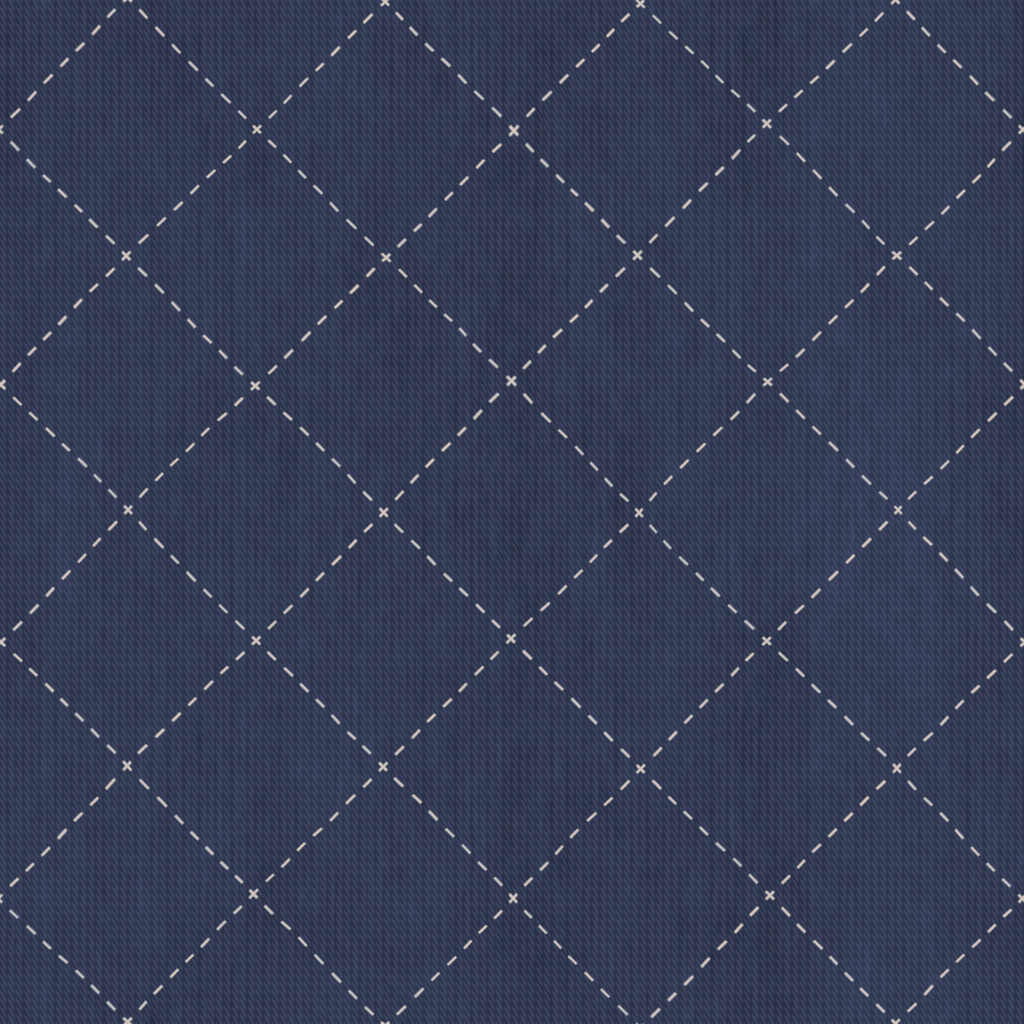} & 
                                    \includegraphics[width=0.16\textwidth,height=0.12\textwidth]{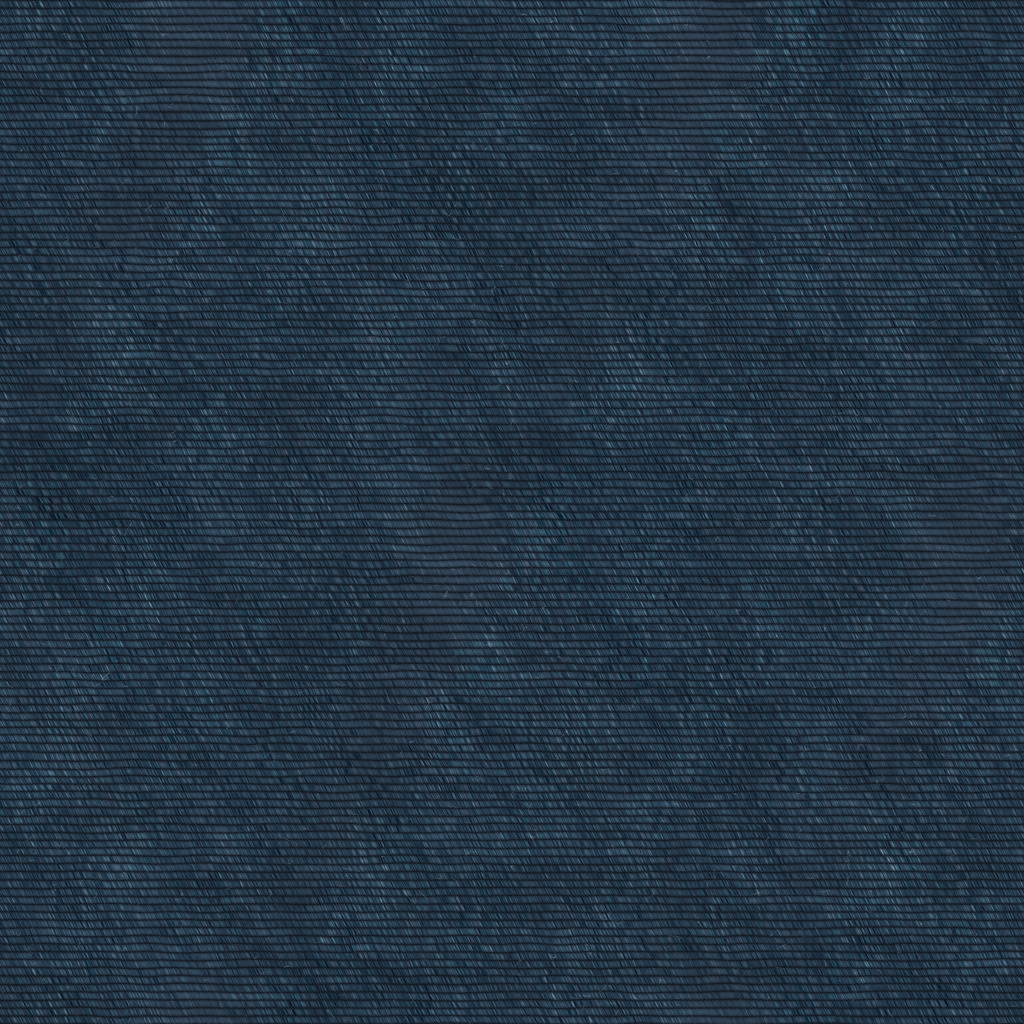} & 
                                    \includegraphics[width=0.16\textwidth,height=0.12\textwidth]{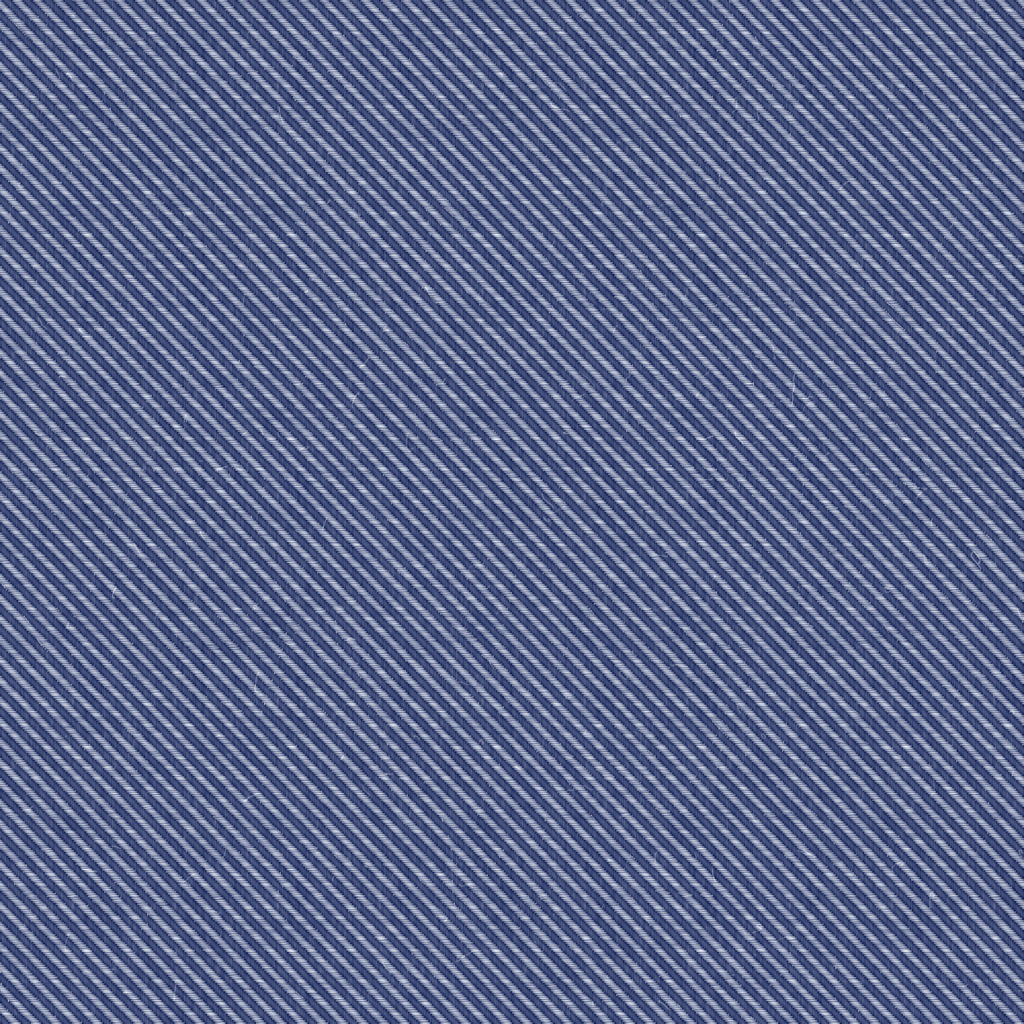} \\
                          
                          & Test  & \includegraphics[width=0.16\textwidth,height=0.12\textwidth]{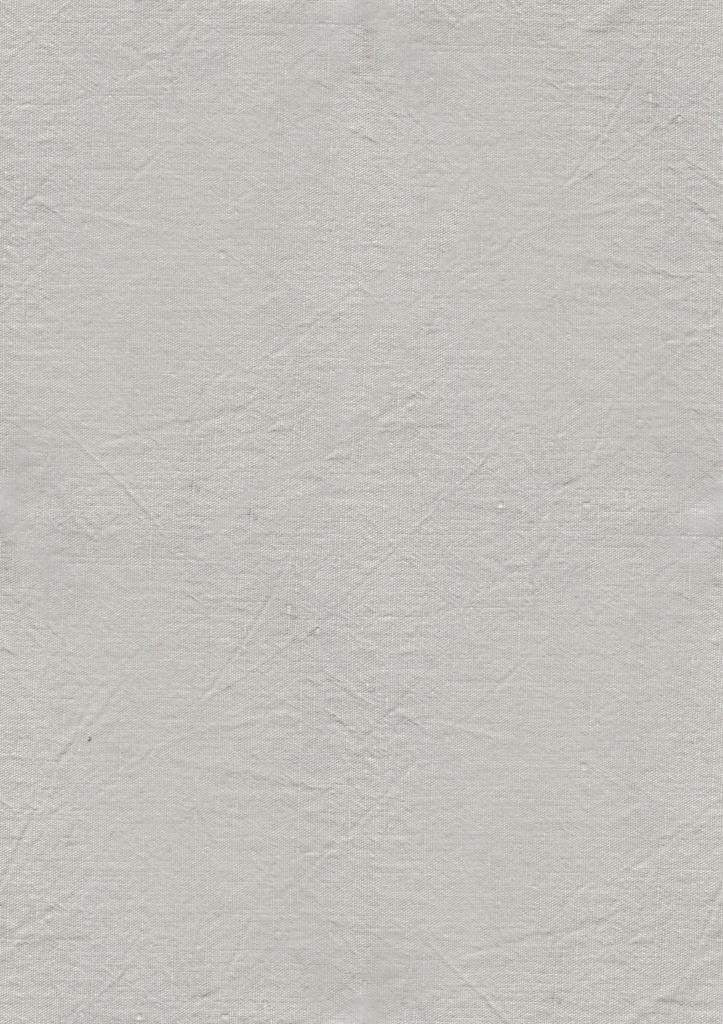} & 
                                    \includegraphics[width=0.16\textwidth,height=0.12\textwidth]{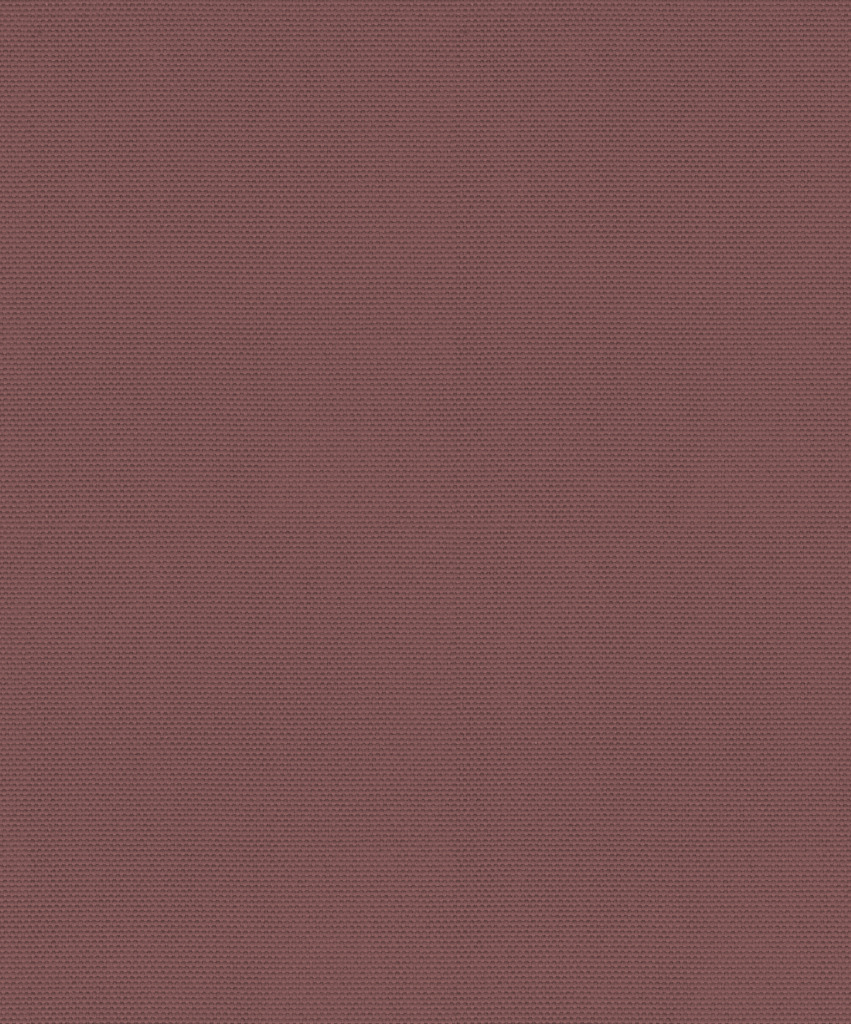} & 
                                    \includegraphics[width=0.16\textwidth,height=0.12\textwidth]{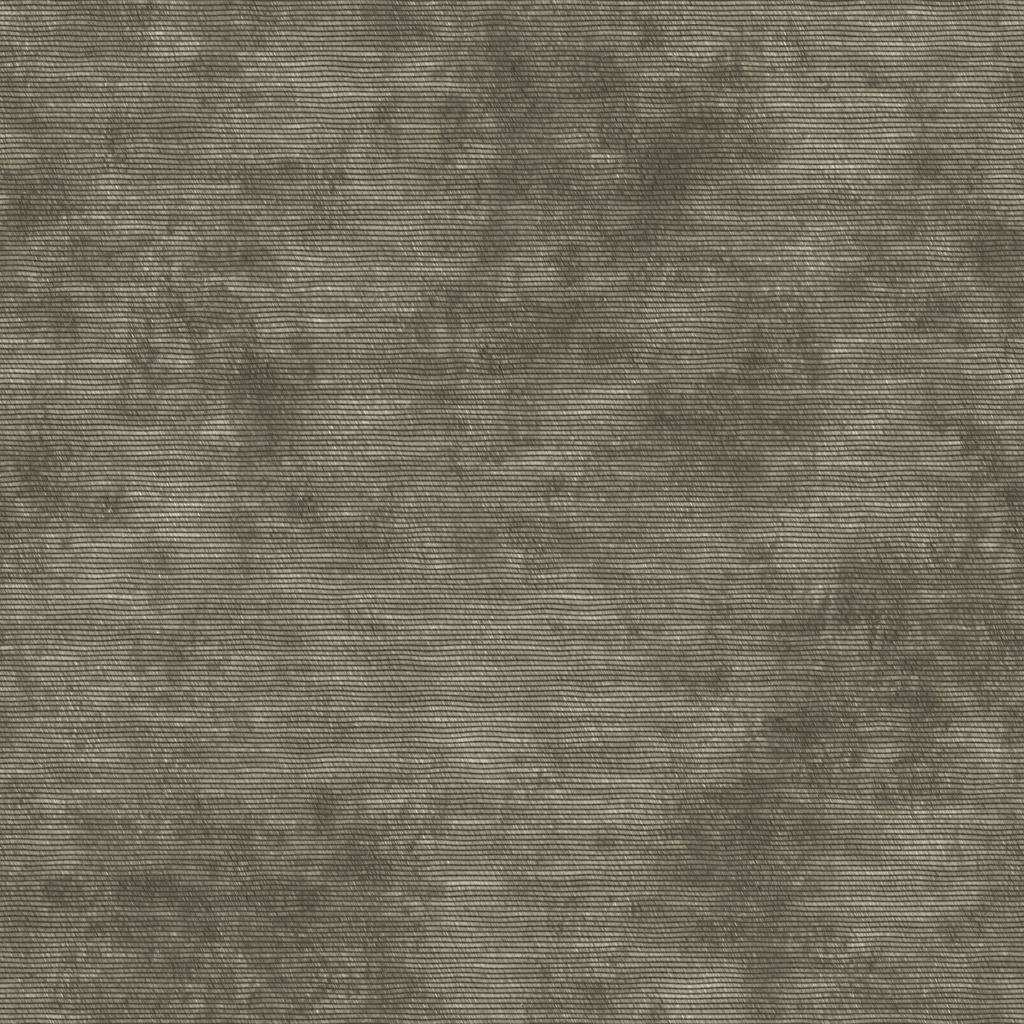} \\
\noalign{\smallskip}
\hline
\noalign{\smallskip}
\multirow{2}{*}{planks}   & Train & \includegraphics[width=0.16\textwidth,height=0.12\textwidth]{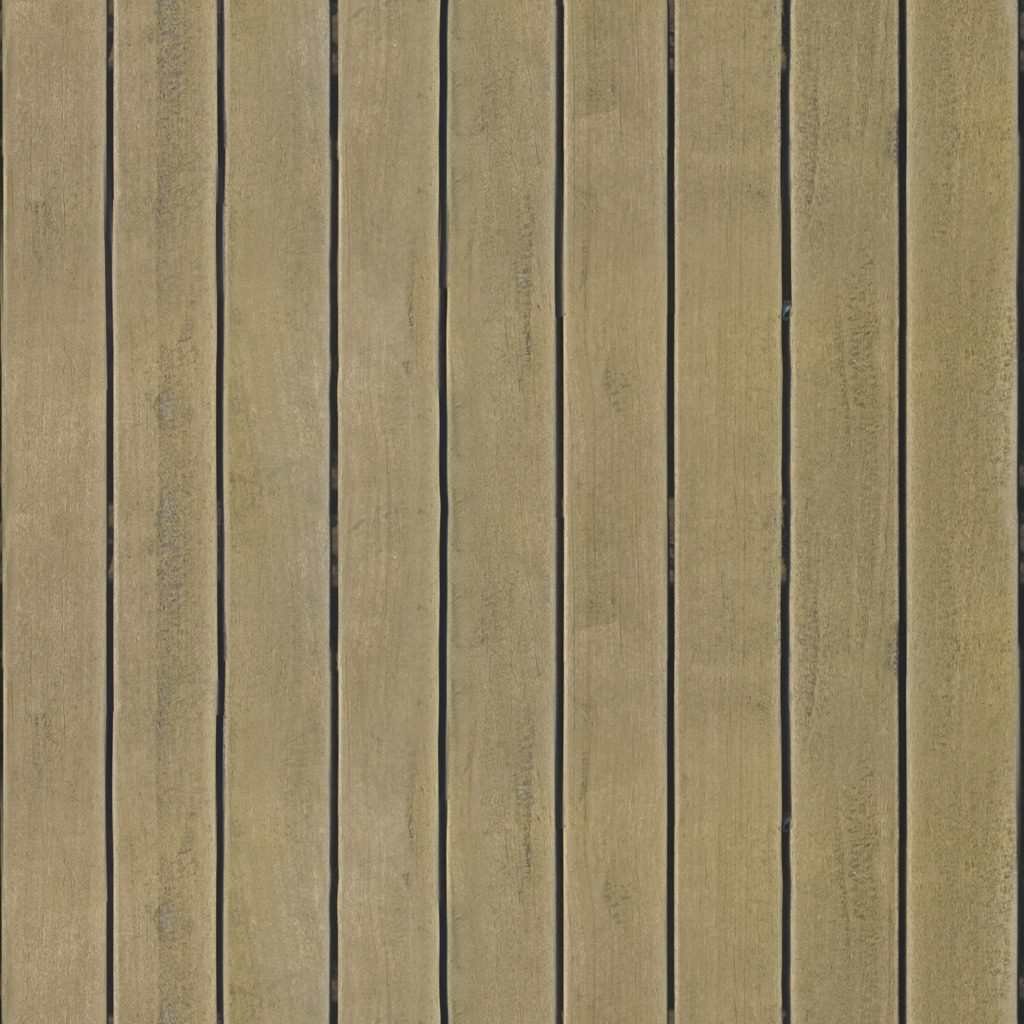} & 
                                    \includegraphics[width=0.16\textwidth,height=0.12\textwidth]{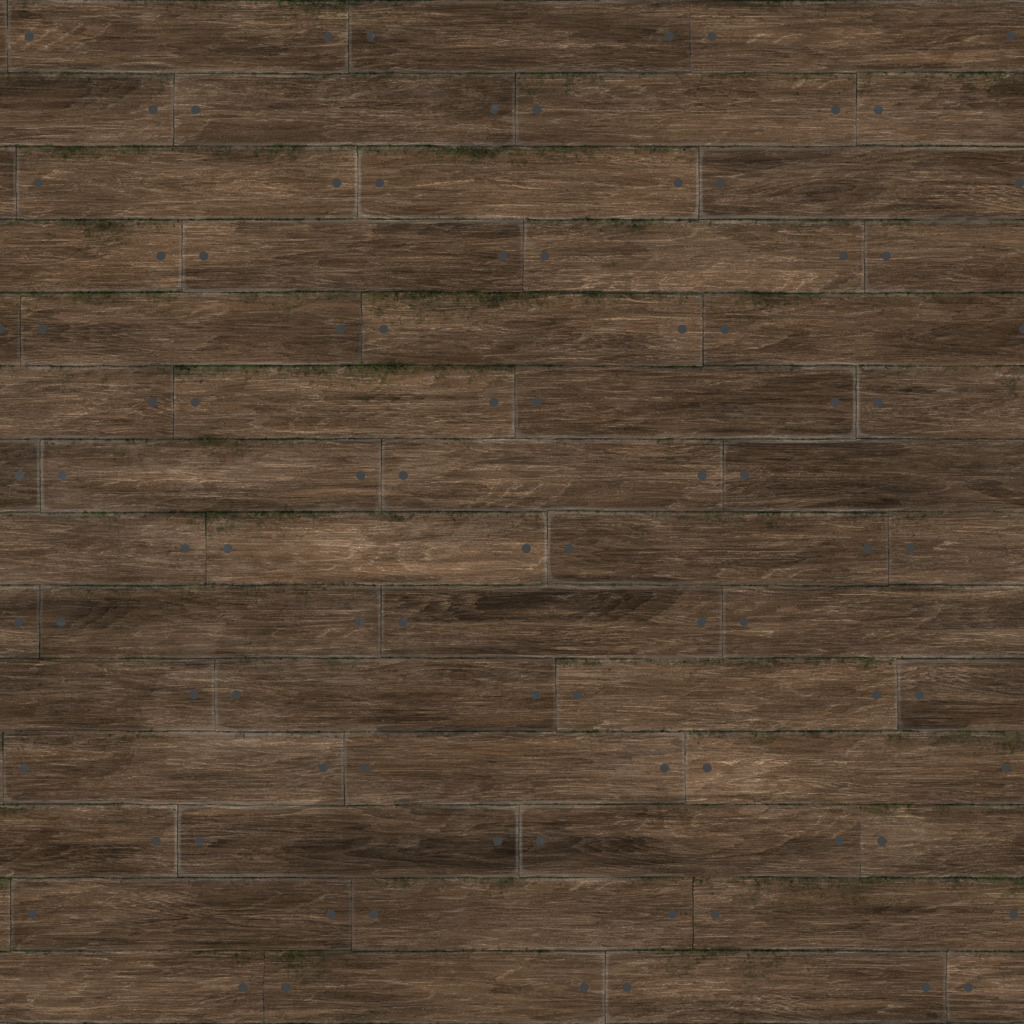} & 
                                    \includegraphics[width=0.16\textwidth,height=0.12\textwidth]{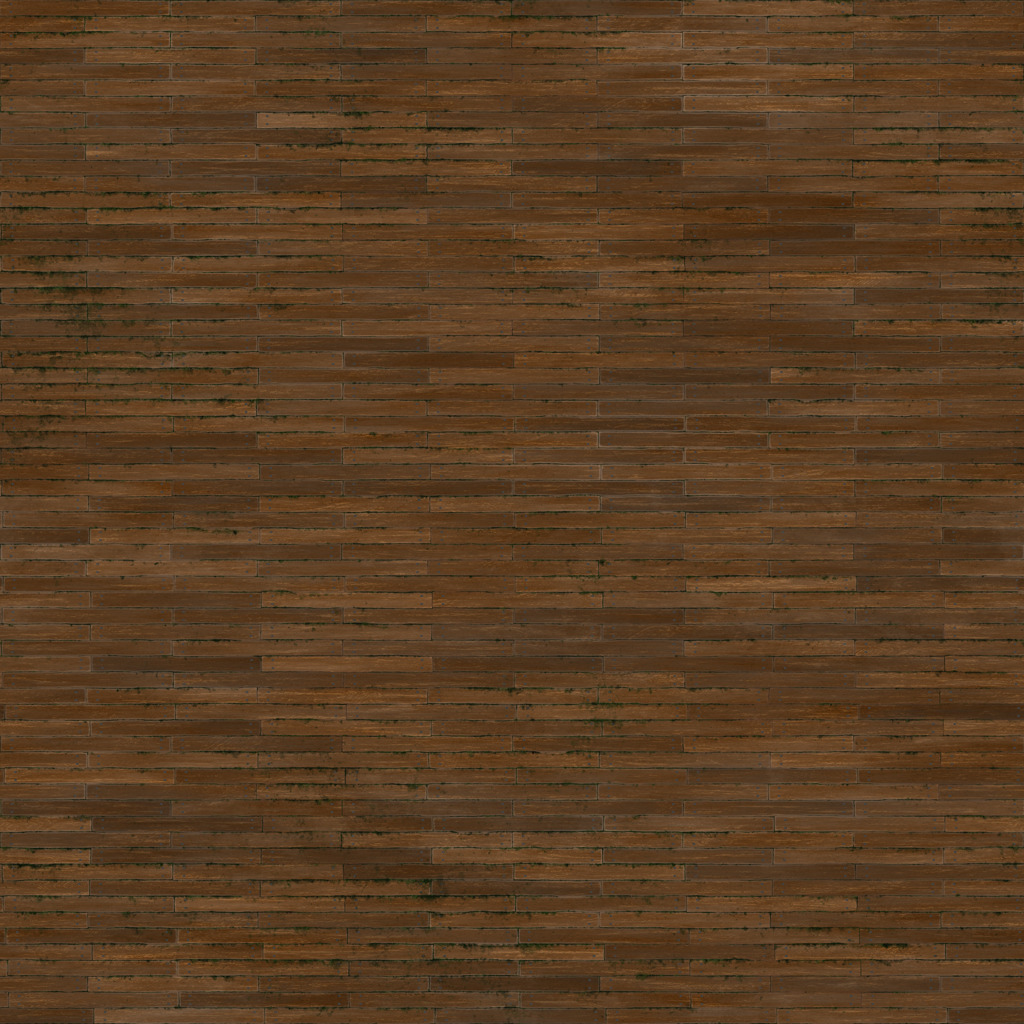} \\
                          
                          & Test  & \includegraphics[width=0.16\textwidth,height=0.12\textwidth]{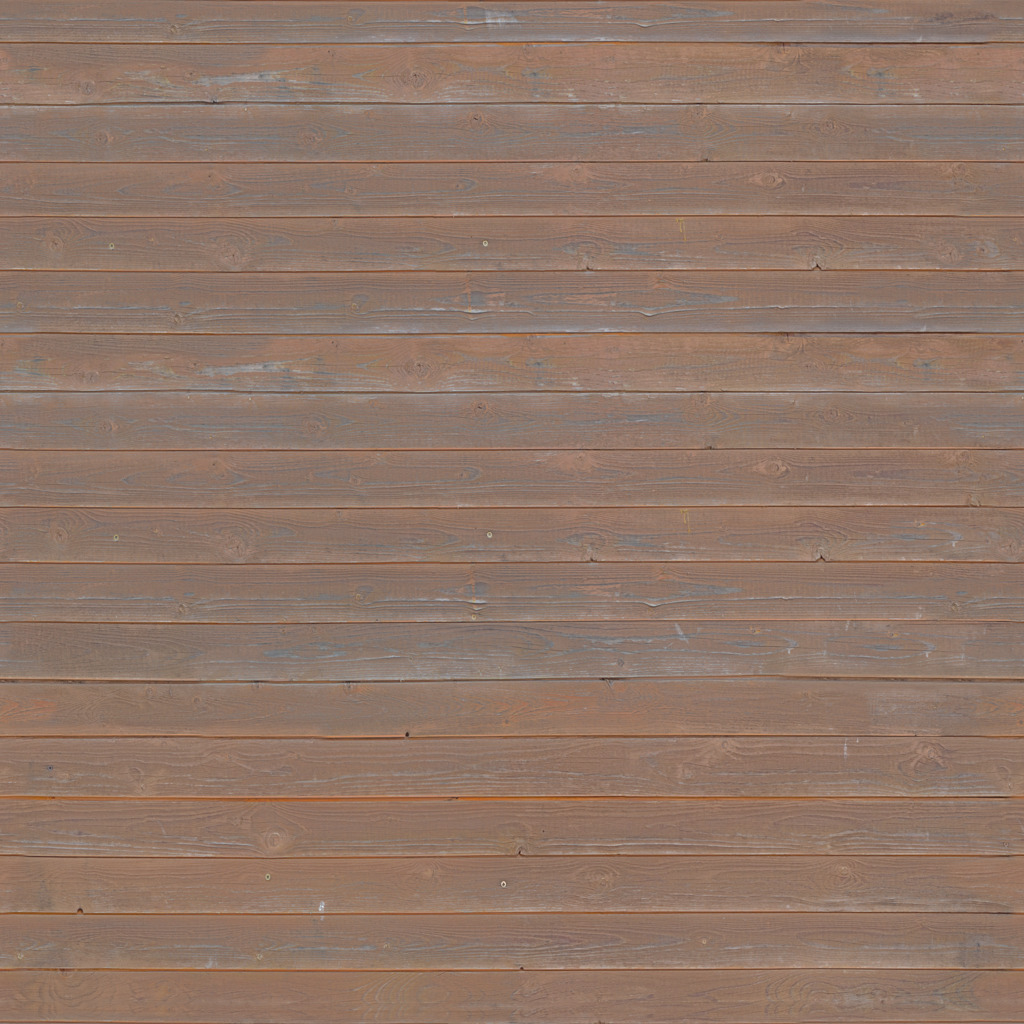} & 
                                    \includegraphics[width=0.16\textwidth,height=0.12\textwidth]{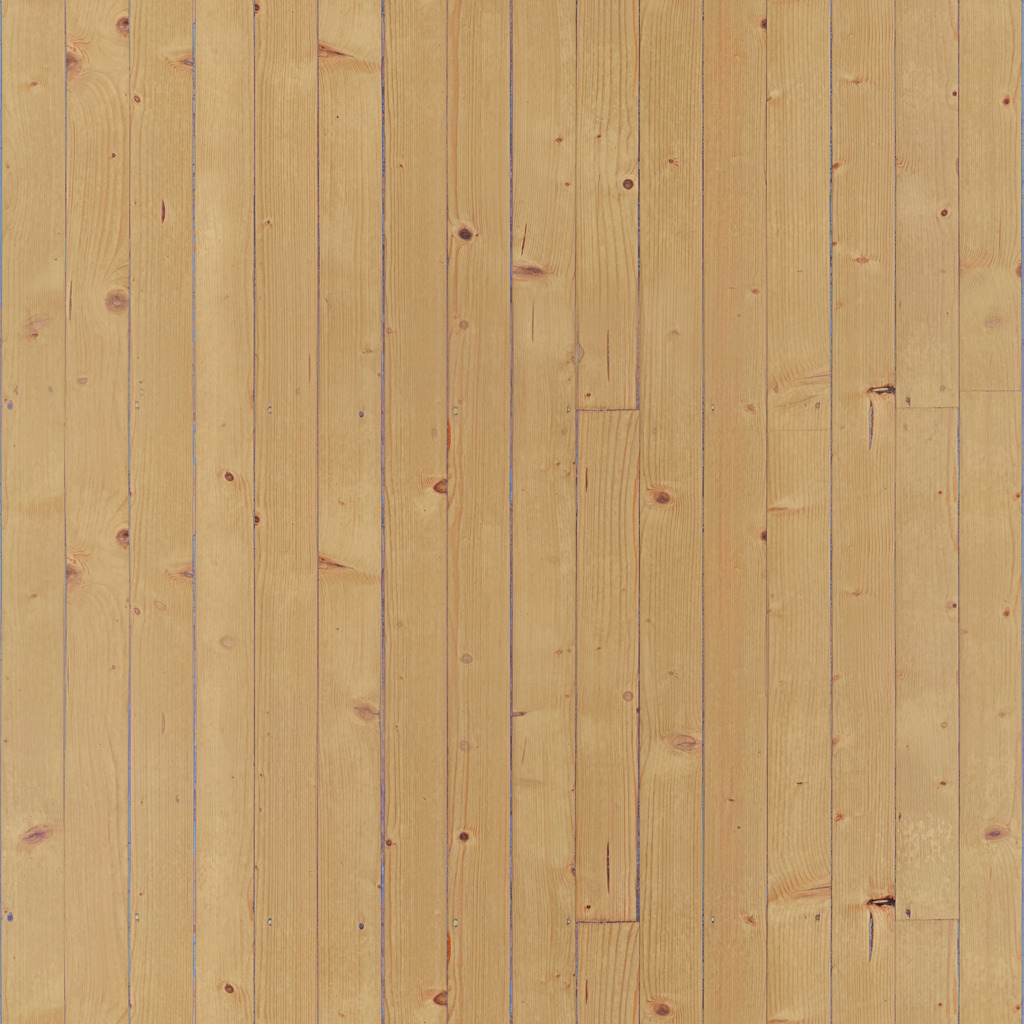} & 
                                    \includegraphics[width=0.16\textwidth,height=0.12\textwidth]{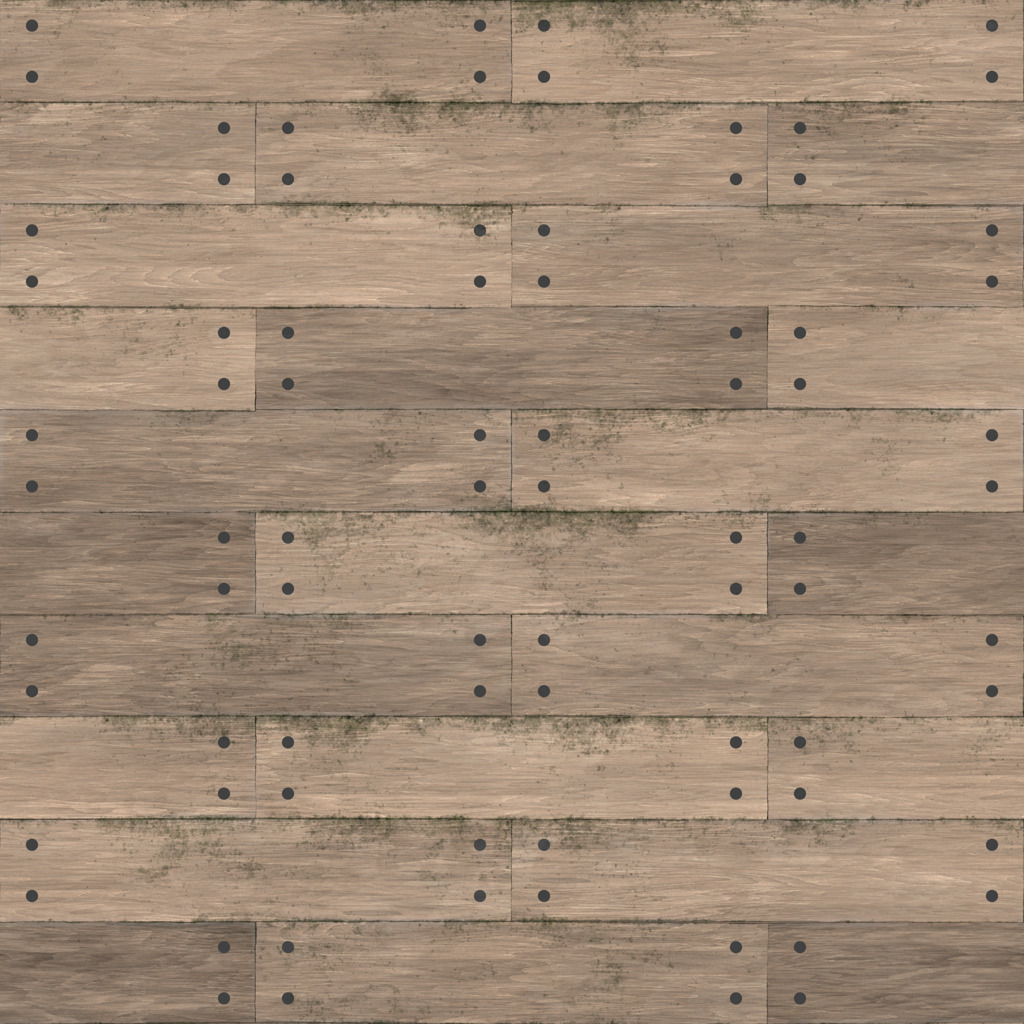} \\
\noalign{\smallskip}
\hline
\noalign{\smallskip}
\multirow{2}{*}{wood}     & Train & \includegraphics[width=0.16\textwidth,height=0.12\textwidth]{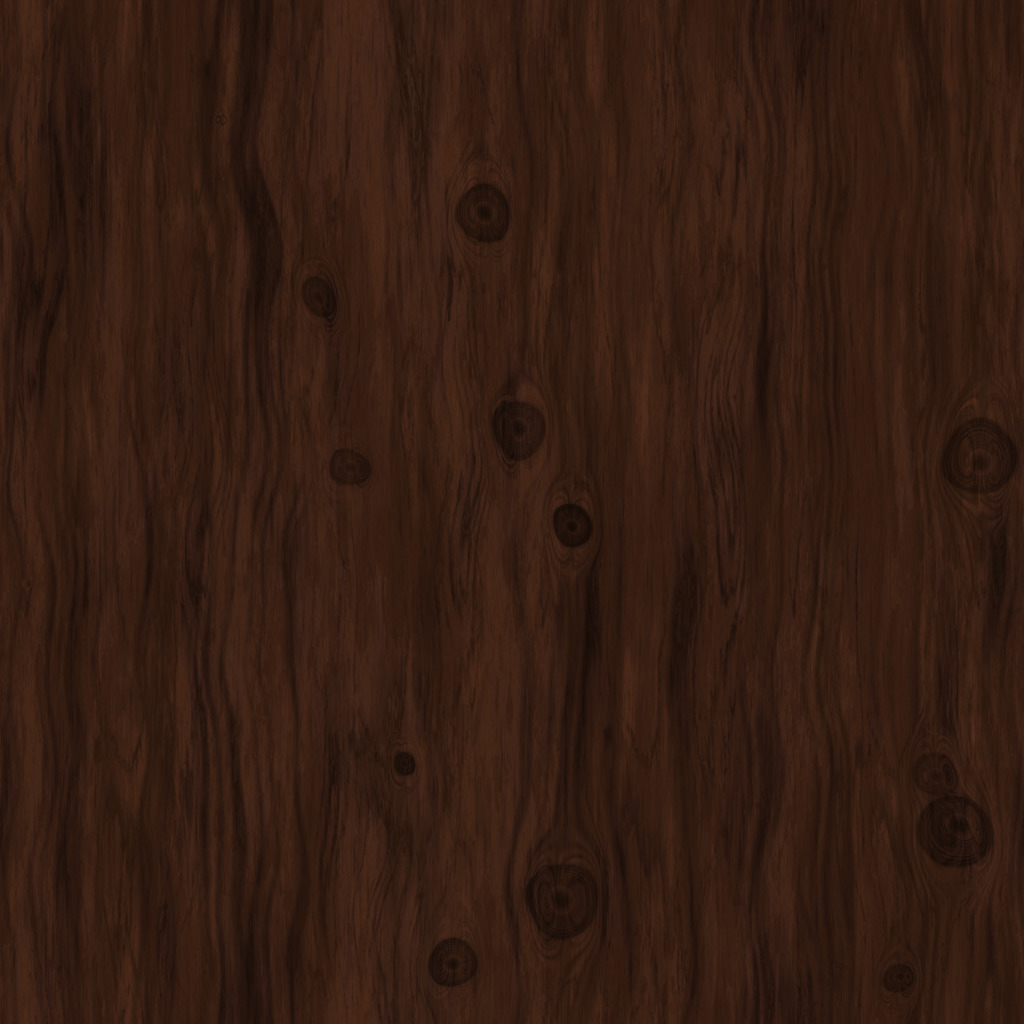} & 
                                    \includegraphics[width=0.16\textwidth,height=0.12\textwidth]{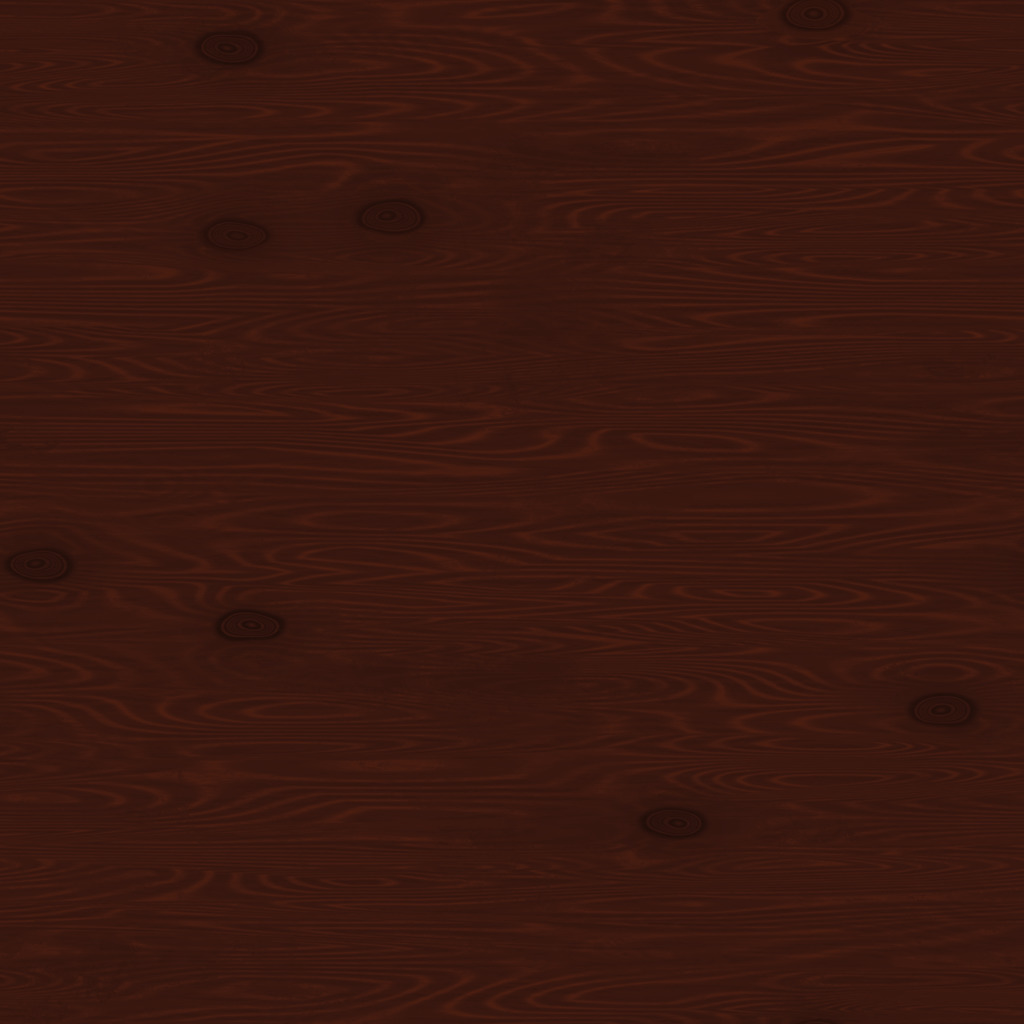} & 
                                    \includegraphics[width=0.16\textwidth,height=0.12\textwidth]{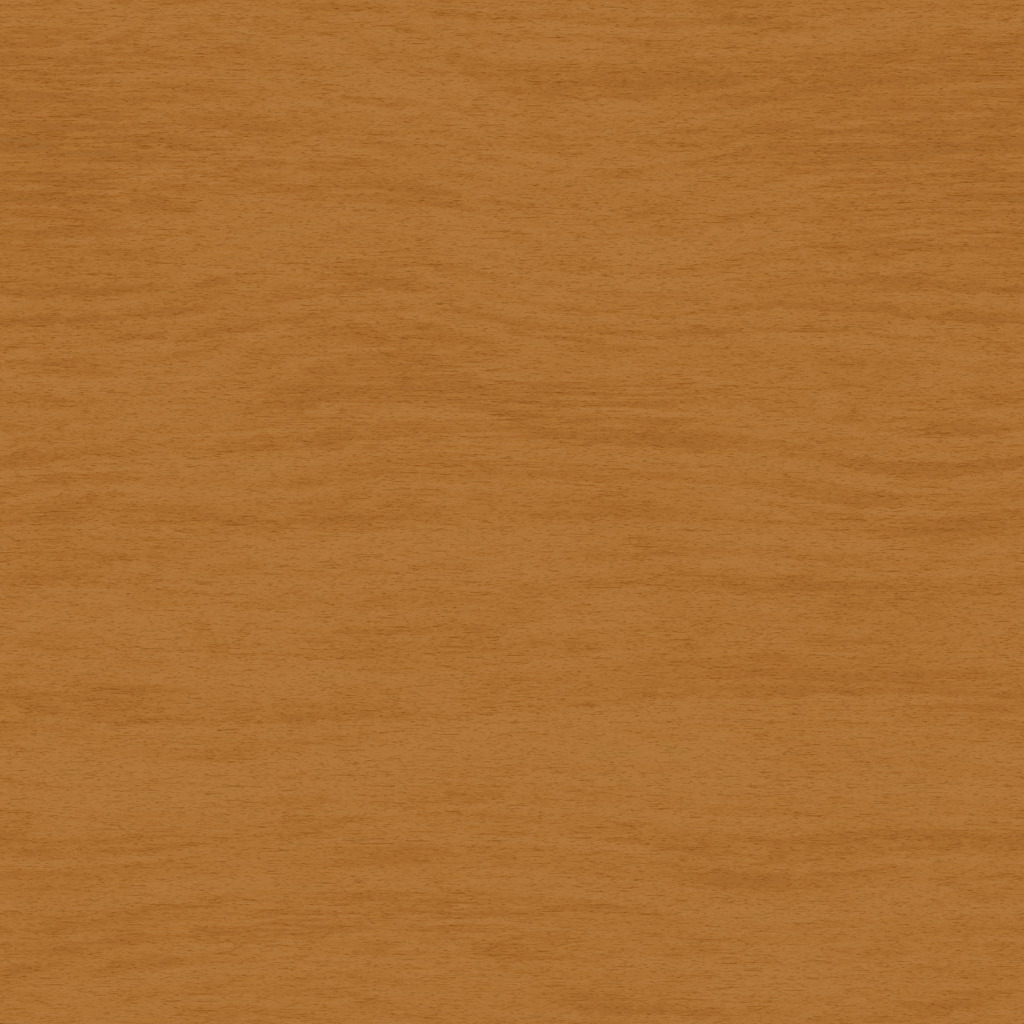} \\
                          
                          & Test  & \includegraphics[width=0.16\textwidth,height=0.12\textwidth]{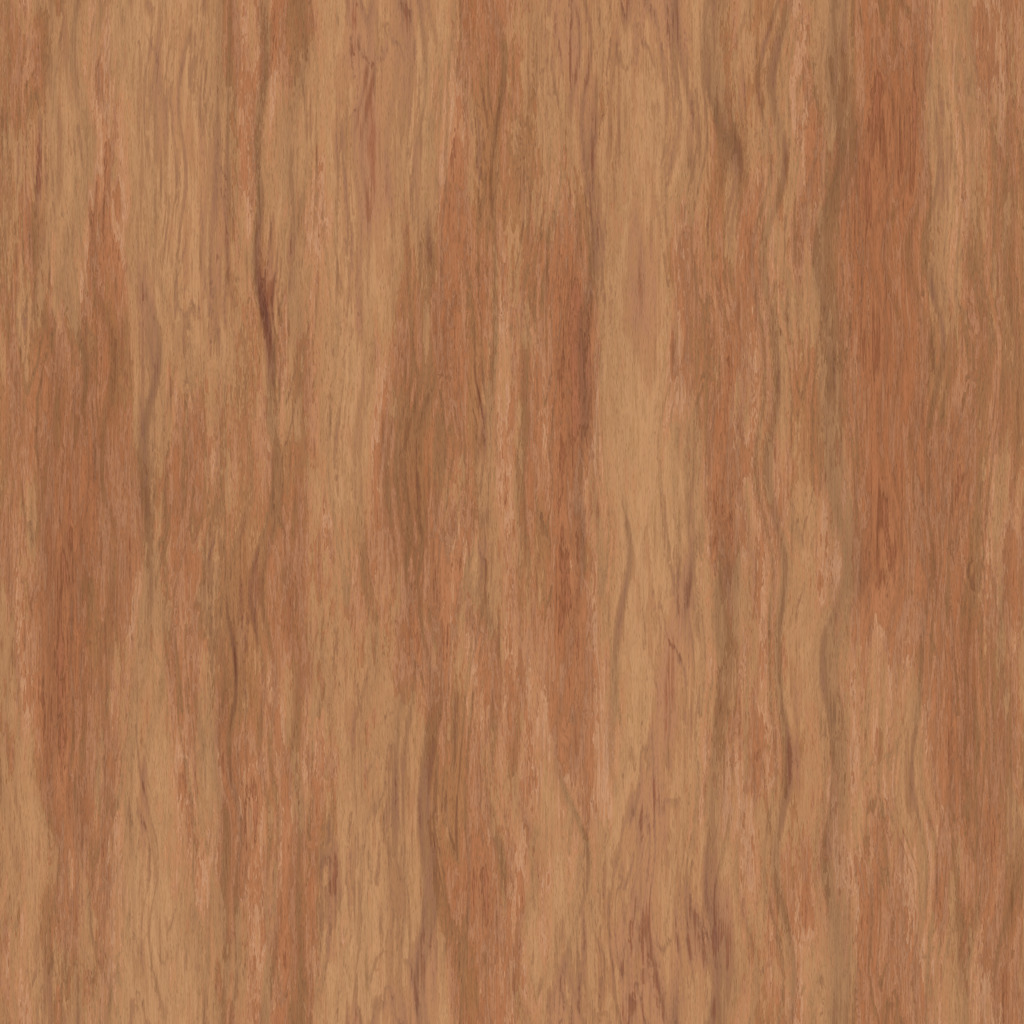} & 
                                    \includegraphics[width=0.16\textwidth,height=0.12\textwidth]{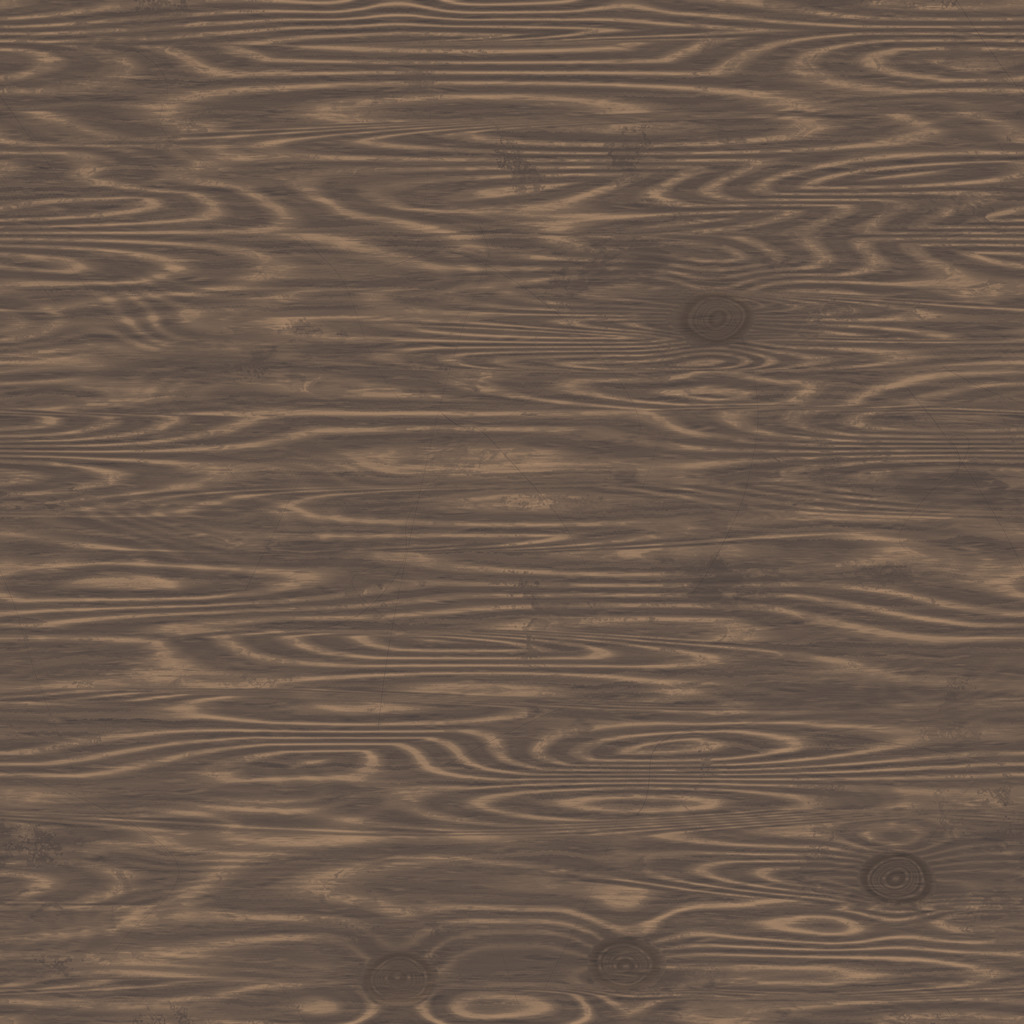} & 
                                    \includegraphics[width=0.16\textwidth,height=0.12\textwidth]{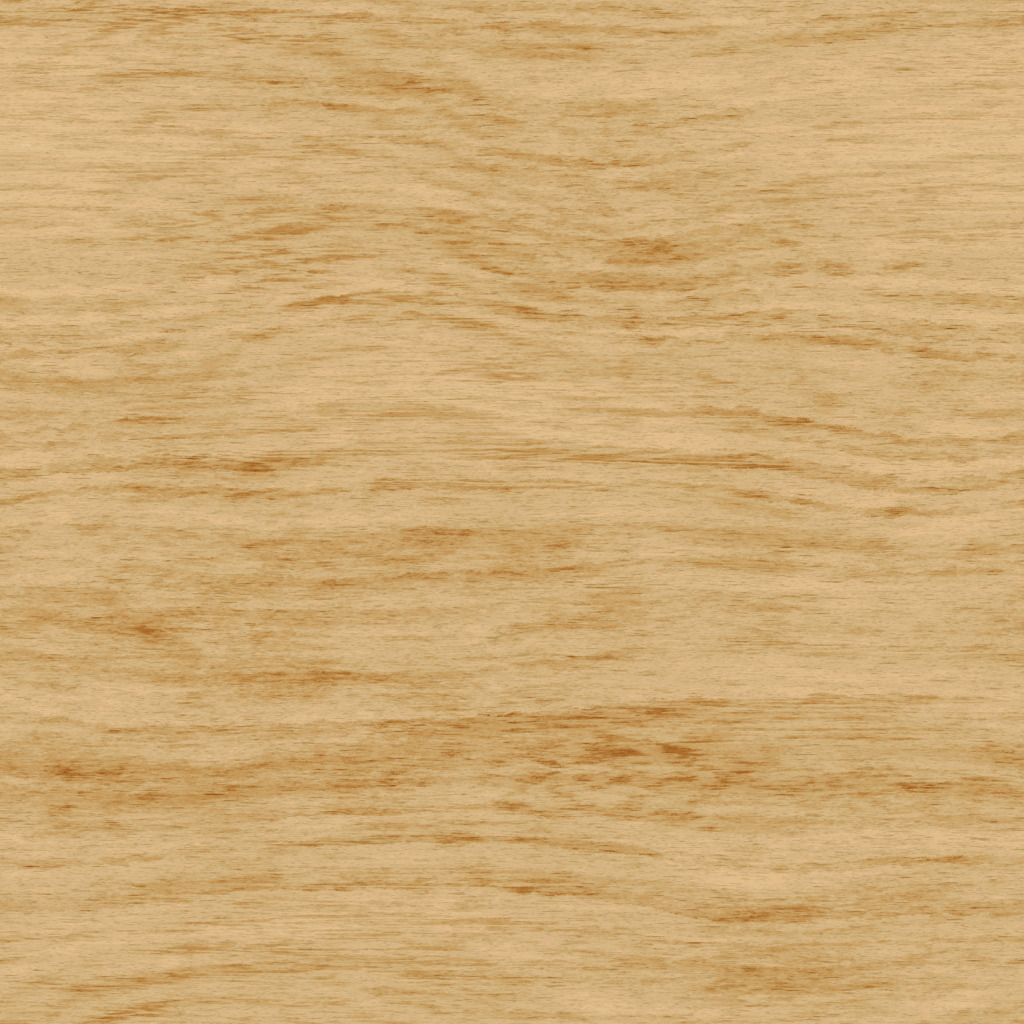} \\
\noalign{\smallskip}
\hline
\noalign{\smallskip}
\end{tabular}
\end{center}
\caption{Examples of training and testing phase textures split from the iGibson 1.0 environment.}
\label{tbl:held-out-textures-examples}
\end{minipage}
\end{table}

\end{document}